\documentclass[journal]{IEEEtran}
\usepackage{graphicx}
\usepackage{amsmath}
\usepackage{booktabs}
\usepackage{bbding}
\usepackage{multirow}
\usepackage{diagbox}
\usepackage{amssymb}
\usepackage{subfigure}
\usepackage{color}
\usepackage{verbatim}
\ifCLASSINFOpdf
\else
\fi
\begin{document}
%
\title{Deep Learning based Joint Geometry and Attribute Up-sampling for Large-Scale Colored Point Clouds}
%
%
%
\author{Yun~Zhang,~\IEEEmembership{Senior~Member,~IEEE}, Feifan Chen, Na Li, Zhiwei Guo, Xu Wang,~\IEEEmembership{Member, IEEE}, \\
Fen Miao, Sam Kwong,~\IEEEmembership{Fellow, IEEE}
 \thanks{Manuscript received xxx; revised xxx.This work was supported by in part by the National Natural Science Foundation of China under Grant 62172400, Shenzhen Key Science and Technology Program under Grant JCYJ20241202124415021, Shenzhen-Hong Kong Collaborative Project Tier-A under Grant SGDX2024011505505010, and in part by Shenzhen Natural Science Foundation under Grant JCYJ20240813180503005.}
\thanks{Y. Zhang and Z. Guo are with the School of Electronics and Communication Engineering, Sun Yat-sen University, Shenzhen Campus, Shenzhen 518107, China (e-mail: zhangyun2@mail.sysu.edu.cn, guozhw7@mail2.sysu.edu.cn).}
\thanks{F. Chen and N. Li are with the Shenzhen Institute of Advanced Technology, Chinese Academy of Sciences, Shenzhen 518055, China (e-mail: \{ff.chen, na.li1\}@siat.ac.cn).}
\thanks{X. Wang is with the College of Computer Science and Software Engineering, Shenzhen University, Shenzhen 518060, China (e-mail: wangxu@szu.edu.cn).}
\thanks{F. Miao is with the Shenzhen Institute for Advanced Study, University of Electronic Science and Technology of China, Shenzhen 518110, China (email:fenmiao@uestc.edu.cn).}
\thanks{S. Kwong is with the Department of Computing and Decision Sciences, Lingnan University, Hong Kong, China (e-mail: samkwong@ln.edu.hk).
}
}
%
%

\markboth{Submitted to IEEE Transactions on Image Processing, Jan~2024}%
{Zhang \MakeLowercase{\textit{et al.}}: Deep Learning based Joint Geometry and Attribute Up-sampling for Large-Scale Colored Point Clouds}
%


\maketitle
\begin{abstract}
Colored point cloud comprising geometry and attribute components is one of the mainstream representations enabling realistic and immersive 3D applications.
To generate large-scale and denser colored point clouds, we propose a deep learning-based Joint Geometry and Attribute Up-sampling (JGAU) method, which learns to model both geometry and attribute patterns and leverages the spatial attribute correlation.
Firstly, we establish and release a large-scale dataset for colored point cloud up-sampling, named SYSU-PCUD, which has 121 large-scale colored point clouds with diverse geometry and attribute complexities in six categories and four sampling rates.
Secondly, to improve the quality of up-sampled point clouds, we propose a deep learning-based JGAU framework to up-sample the geometry and attribute jointly. It consists of a geometry up-sampling network and an attribute up-sampling network, where the latter leverages the up-sampled auxiliary geometry to model neighborhood correlations of the attributes.
Thirdly, we propose two coarse attribute up-sampling methods, Geometric Distance Weighted Attribute Interpolation (GDWAI) and Deep Learning-based Attribute Interpolation (DLAI), to generate coarsely up-sampled attributes for each point. Then, we propose an attribute enhancement module to refine the up-sampled attributes and generate high quality point clouds by further exploiting intrinsic attribute and geometry patterns.
Extensive experiments show that Peak Signal-to-Noise Ratio (PSNR) achieved by the proposed JGAU are 33.90 dB, 32.10 dB, 31.10 dB, and 30.39 dB when up-sampling rates are 4$\times$, 8$\times$, 12$\times$, and 16$\times$, respectively. Compared to the state-of-the-art schemes, the JGAU achieves an average of 2.32 dB, 2.47 dB, 2.28 dB and 2.11 dB PSNR gains at four up-sampling rates, respectively, which are significant.
\end{abstract}

\begin{IEEEkeywords}
Large-scale colored point cloud, joint geometry and attribute up-sampling, deep learning.
\end{IEEEkeywords}

%
\IEEEpeerreviewmaketitle

\section{Introduction}
%
%
%
%
\IEEEPARstart
{P}{oint} cloud is a set of discrete and unstructured points representing the geometry in Three Dimensional (3D) space (x,y,z) and attributes (e.g., RGB color, transparency, reflectance) of 3D scenes, which enables immersive 3D depth perception and unique Six Degrees of Freedom (6DoF) view interactions \cite{[26]}. Nowadays, the point cloud is developing towards larger scale, colorful and higher density, which has been one of the main representations for emerging cutting-edge 3D applications, such as smart city, autonomous vehicle, 3D manufacturing, robotic, Augmented Reality (AR) and Virtual Reality (VR) \cite{PCQA}. However, due to the limitations of resolution, accuracy and computational capability in acquisition devices and 3D modelling algorithms, the captured raw point clouds are usually sparse, which hinders realistic 3D applications, such as AR/VR and remote control. To handle this problem, point cloud enhancement is highly demanded for capturing large scale, high quality and dense point clouds.

\subsection{Related Work}
Point cloud up-sampling is one of the key solutions for point cloud enhancement, which generates a set of denser points from sparse ones. Since geometry components stand for the geometrical coordinates of a point, a geometry up-sampling is to interpolate more point coordinates in 3D space. Moreover, since there are attribute components attached to each point in colored point clouds, the attributes shall be also up-sampled and assigned to the geometrically interpolated 3D points, which is an attribute up-sampling. To enable more realistic 3D applications, many researches have been devoted to the point cloud up-sampling schemes, which are categorized as geometry and attribute up-sampling schemes.

\subsubsection{Point Cloud Geometry Up-sampling}
Conventionally, geometry up-sampling methods were optimized with empirical priors and hand-crafted features.
Lipman \textit{et al}. \cite{[14]} proposed a parametrization-free method for both point re-sampling and surface reconstruction based on the Locally Optimal Projection (LOP) operator. Subsequently, Huang \textit{et al}. \cite{[15]} proposed a weighted LOP to iteratively consolidate point clouds by using normal estimation, which was robust to noise and outliers.
However, the LOP-based methods assumed that points were sampled from smooth surfaces, which degraded the up-sampling performance of complex structure areas, e.g., sharp edges and corners.
Huang \textit{et al}. \cite{[16]} proposed an edge-aware up-sampling method that used reliable normal vectors to preserve sharp edges. 
Wu \textit{et al}. \cite{[17]} proposed a depth point-based point set representation method for point cloud completion.
However, point clouds with noisy points and outliers were not considered. Moreover, limited hand-crafted features reduce the accuracy and applicability of the up-sampling methods.

Recently, many deep learning-based point cloud up-sampling methods have been proposed to mine pattern knowledge from data.
Qi \emph{et al.} \cite{[2]} proposed a deep learning based point cloud network, called PointNet, where shared Multi-Layer Perceptron (MLP) and symmetric maximum pooling operation were used to extract irregular and unordered features of point clouds. To exploit global and local information more effectively in extracting geometry features, PointNet++ \cite{[18]} was proposed by aggregating neighborhood points. Li \textit{et al.} \cite{[19]} proposed a PointCNN for up-sampling, where irregular point clouds were transformed to a regularized 2D representation for convolution-based feature extraction.
Yu \textit{et al}. \cite{[20]} proposed a Point cloud geometry Up-sampling Network (PU-Net) based on the PointNet. However, the PU-Net overlooked the local geometry.
Qian \textit{et al}. proposed a Point cloud Up-sampling via Geometry-centric Network (PUGeo-Net) \cite{[5]} and Magnification-Flexible Up-sampling (MFU) \cite{[6]} by using local geometry difference constraints.
Wang \textit{et al}. \cite{[22]} proposed a patch-based progressive 3D point set up-sampling, where a progressive up-sampling factor was used to achieve good performance on large up-sampling factors. However, the supported up-sampling factor shall be the power of 2. {To up-sample a point cloud to an arbitrary-scale, Ye \textit{et al}. \cite{[25]} proposed an arbitrary scale up-sampling network Meta-PU, where the point cloud was up-sampled to a super-high resolution first and then down-sampled to the arbitrary target scale by using Farthest Point Sampling (FPS). Zhang \textit{et al}. \cite{[48]} proposed a novel progressive point cloud up-sampling framework to solve the non-uniform distribution problem during point cloud up-sampling.}

{To further improve the performance of geometry up-sampling, more advanced neural networks and learning strategies were investigated.} Li \textit{et al}. \cite{[4]} developed a Generative Adversarial Network (GAN) for the point cloud up-sampling, named PU-GAN, using a composite loss function constraint to generate a uniformly distributed set of up-sampled points. Hao \textit{et al}. \cite{PUFATIP2023} developed a PUFA-GAN scheme by further considering geometry frequency in training GAN based geometry up-sampling.
Qian \textit{et al}. \cite{[23]} introduced graph convolution to geometry up-sampling and proposed a multi-scale graph convolution network to improve the performance of geometry.
Li \textit{et al}. \cite{[24]} proposed a Disentangled refinement for Point cloud Up-sampling (Dis-PU) by decomposing point cloud geometry up-sampling into a two-stage task. It generated a coarse point set which was then enhanced with local and global refinement units.
{Liu \textit{et al}. \cite{PU-Mask} proposed a mask-guided transformer-style asymmetric autoencoder for geometry up-sampling, where local geometry distance based virtual masks were generated and used as constraints for feature restoration and rectification.
Mao \textit{et al}. \cite{PU-Flow} proposed a point cloud geometry up-sampling network, {named PU-flow}, by learning the local coordinate interpolation in a latent space. It incorporated normalizing flows and weight estimation to produce uniformly distributed dense points.}
Feng \textit{et al}. \cite{[7]} learned neural fields to represent high-resolution geometric shapes and surfaces, which were down-sampled to generate arbitrary scale point clouds.
In these works, advances in deep neural networks significantly improve the geometry up-sampling. However, {they were developed to up-sample the small-scale and sparse point clouds with a limited number of points, which have not exploited the properties of large-scale and dense point clouds. Moreover, they only up-sampled the geometric coordinates of point clouds to increase the density of points, which were not applicable to the attributes.}

To handle large-scale point clouds upsampling for semantic recognition, Chen \textit{et al}. \cite{ChenTMM23} proposed a structure semantics guided LiDAR up-sampling network for indoor robotics, called SGSR-Net, where Coordinate Attention with Squeeze and Excitation (CASE) was used for feature extraction and structure semantics were exploited in the geometry up-sampling. The LiDAR point cloud is developed for large and sparse scenarios. In addition to conventional geometry quality, Li \textit{et al}. \cite{LiTMM22} proposed a point cloud up-sampling to improve the semantic classification of sparse point cloud.
{They aimed to improve the semantic recognition performance, which differed from the objective of improving the visual quality for human vision.}

\begin{figure*}[t]
\centering
\subfigure[]{
\includegraphics[width=0.48\textwidth]{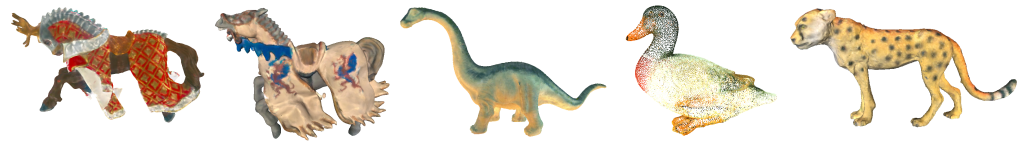}
}
\subfigure[]{
\includegraphics[width=0.48\textwidth]{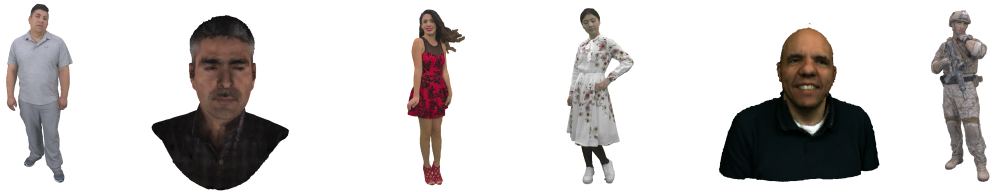}
}
\subfigure[]{
\includegraphics[width=0.48\textwidth]{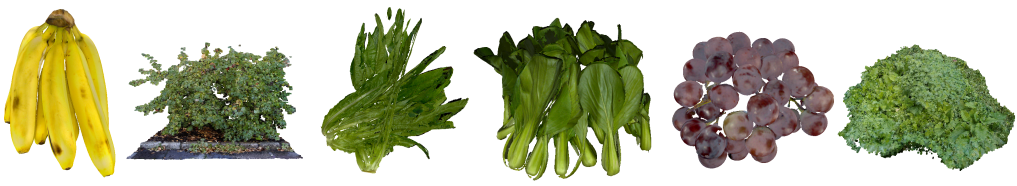}
}
\subfigure[]{
\includegraphics[width=0.48\textwidth]{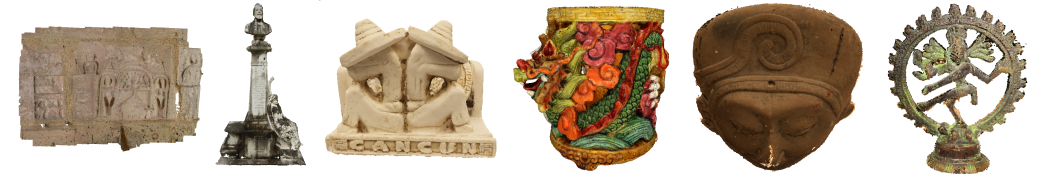}
}
\subfigure[]{
\includegraphics[width=0.48\textwidth]{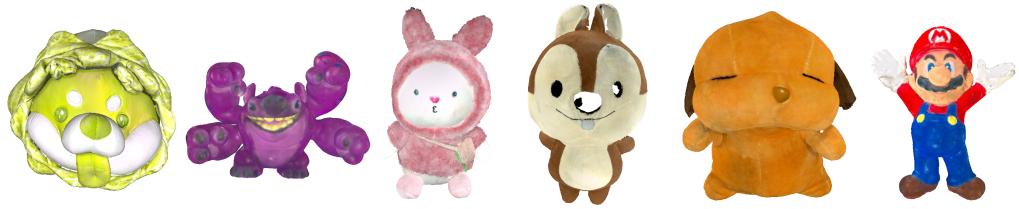}
}
\subfigure[]{
\includegraphics[width=0.48\textwidth]{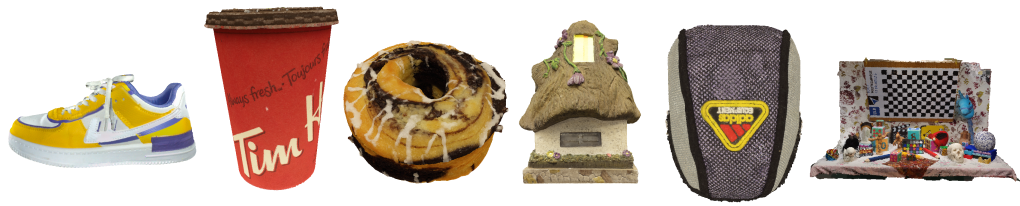}
}
\caption{Visualized samples of the proposed SYSU-PCUD, which has 121 colored point clouds in six categories and four sampling rates (4$\times$, 8$\times$, 12$\times$, and 16$\times$), (a) animal, (b) people, (c) plant, (d) sculpture, (e) toy, (f) others.}
\label{fig4}
\end{figure*}
\subsubsection{Point Cloud Attribute Up-sampling}
With the increasing demands for colored and dense point clouds in realistic 3D applications, point cloud attribute up-sampling is highly desired.
Qian \textit{et al}. \cite{[6]} extended the MFU to attribute up-sampling by exploiting spatial attribute correlation among the nearest neighboring points.
Heimann \textit{et al}. \cite{[38]} proposed a Frequency-Selective Mesh-to-Mesh Resampling (FSMMR) method for point cloud attribute up-sampling, which performed attribute interpolation in a 2D plane before projecting to 3D space. However, the introduced occlusions and geometrical distortions in 2D-to-3D projection may degrade the quality of attribute up-sampling. Also, additional computational overheads and delays were caused. Dinesh \textit{et al}. \cite{[44]} proposed a super-resolution algorithm for colored point clouds by adding interior points, where Fast Graph Total Variation (FGTV) from surface normals and color values were used as loss function in training. Generally, these methods utilized limited empirical or handcrafted features. High dimensional attribute features from data statistics have not been exploited in enhancing the attribute up-sampling.

{To learn attribute pattern knowledge from data, deep learning-based point cloud attribute up-sampling schemes were proposed to enhance the up-sampling quality. Wang \textit{et al}. \cite{[46]} proposed a Color Up-sampling Network (CU-NET) for point cloud attribute, where sparse convolution and neural implicit functions were utilized for attribute prediction.
However, it was only validated on human point clouds, and the quality for larger up-sampling rates could be further improved. By extending the feature embedding of PU-Net, Naik \textit{et al}. \cite{[45]} proposed a deep neural network-based point cloud up-sampling, called FeatureNet, where Earth Mover's Distance (EMD) and MSE losses were used for geometry and attribute, respectively.} However, the geometry and attribute features of point clouds were up-sampled separately. The correlation between geometry and attribute was not fully exploited. {To enhance the quality of compressed point clouds, Guarda \textit{et al}. \cite{IT-DL-PCC} proposed a deep learning based super-resolution as post-processing of deep point cloud compression, named IT-DL-PCC, which treated the point clouds as six-dimensional data and up-sampled the geometry and attribute simultaneously. The inartistic dependencies between geometry and attribute were exploited implicitly with Inception-Residual Block (IRB) based neural network.}
{Overall, these attribute up-sampling methods could be further improved by considering the intrinsic spatial pattern of attribute and the correlation between geometry and attribute. Moreover, these works were learned from a small set of point clouds, which reduced the learning capability and generalization of the learned up-sampling models. A large-scale colored and dense point cloud database is required to enhance the model learning and validation.}

\subsection{Contributions of This Work}
In this paper, we propose a deep learning-based Joint Geometry and Attribute Up-sampling (JGAU) method for large-scale colored point clouds. The main contributions are

\begin{itemize}
\item We build a large-scale Point Cloud Up-sampling Dataset (SYSU-PCUD) containing 121 source colored point clouds with diverse contents in six categories, various numbers of points, dynamic geometry and attribute complexities. Meanwhile, each source point cloud has four sparse versions with sampling rates 4$\times$, 8$\times$, 12$\times$, and 16$\times$. To our best knowledge, this is the largest dataset for colored point cloud up-sampling network training and evaluation, {which has been released to publicity \cite{SYSU-PCUD}}.
\item We model the point cloud up-sampling problem as a joint geometry and attribute up-sampling, and propose a deep learning-based JGAU framework, which learns to model geometry and attribute dependencies and to exploit the spatial attribute correlation.
\item We propose an attribute up-sampling network to exploit intrinsic attribute and geometry patterns of point clouds. It includes a coarse attribute up-sampling module to up-sample attributes and an Attribute Enhancement Module (AEM) to refine the coarsely up-sampled attribute.
\end{itemize}

The remainder of this paper is organized as follows. Section II presents the established large-scale dataset for colored point cloud up-sampling. Section III formulates the point cloud up-sampling problem. Section IV presents the proposed JGAU in detail. Experimental results are analyzed in Section V. Finally, conclusions are drawn in Section VI.

\section{Building a Large-Scale Colored Point Cloud Dataset for Up-sampling}
\label{Sec:database}
Most of the existing point clouds are in small-scale, sparse and colorless, such as \cite{[4]}\cite{[5]}, which are mainly used for geometry up-sampling. However, a point cloud consists of geometry representing the coordinates of a point and attribute representing physical attributes of the point, such as color and transparency. Due to their different properties, the attribute up-sampling totally differs from the geometry up-sampling. Moreover, large-scale point clouds have more complicated geometry structure and attribute details as compared with the small-scale ones. To develop and analyze the colored point cloud up-sampling schemes, we built a large-scale colored point cloud dataset consisting of 121 source colored point clouds. 43 source colored point clouds were captured by 3D Scanner EinScan Pro 2X 2020. The rest 78 were collected from SIAT-PCQD \cite{[26]}, Moving Picture Experts Group (MPEG) point cloud \cite{[27]}, Greyc 3D colored mesh database \cite{[28],PCQA}, which were initially developed for quality assessment \cite{[26]}, coding \cite{ZhangTIP2023,TSC-PCAC} and mesh up-sampling \cite{[38]}.

\begin{figure}[t]
	\centering
	\subfigure[]{
		\begin{minipage}[t]{1.0\linewidth}
			\centering
			\includegraphics[width=0.9\textwidth]{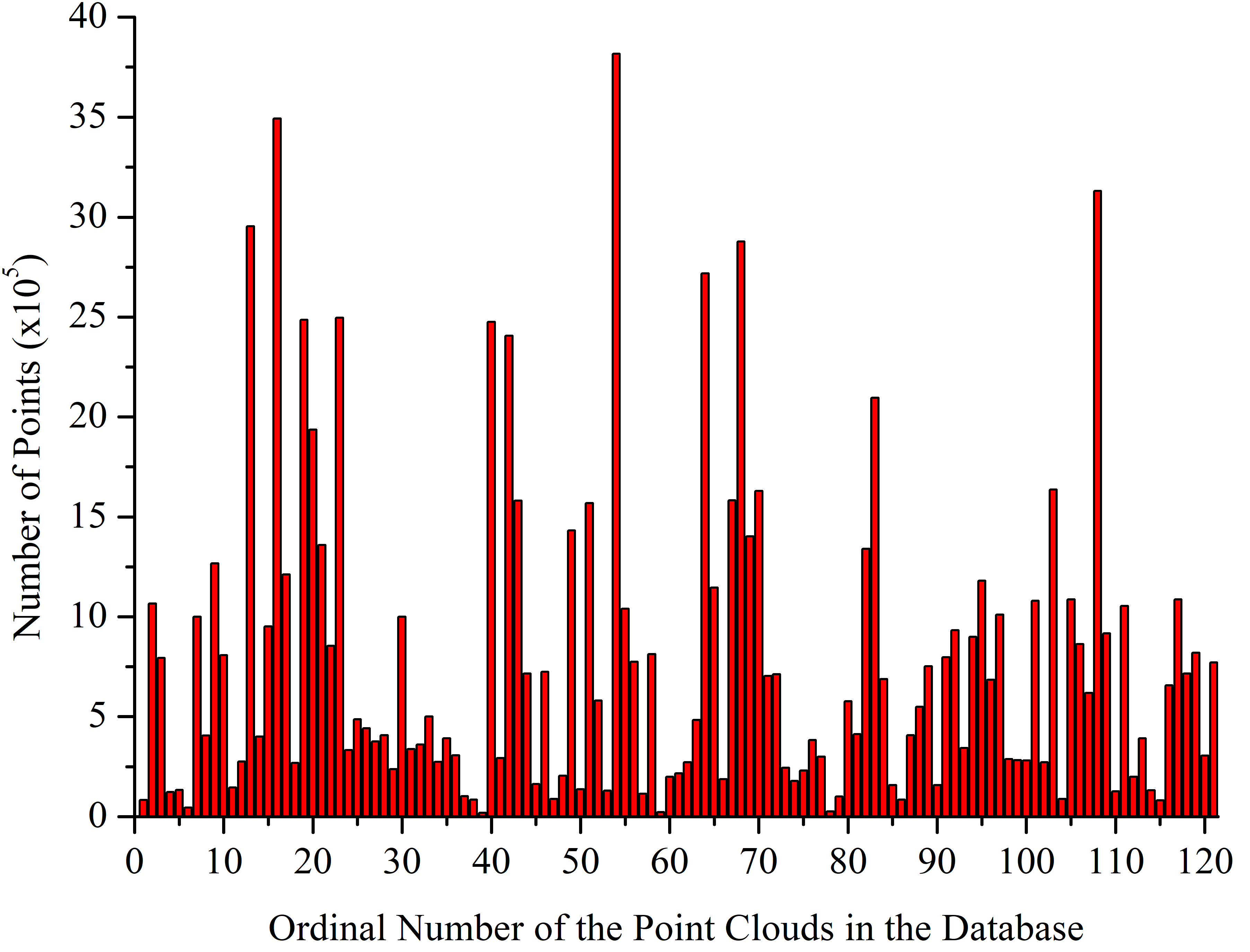}
		\end{minipage}
\label{fig7}
	}
	\subfigure[]{
		\begin{minipage}[t]{1.0\linewidth}
			\centering
			\includegraphics[width=0.9\textwidth]{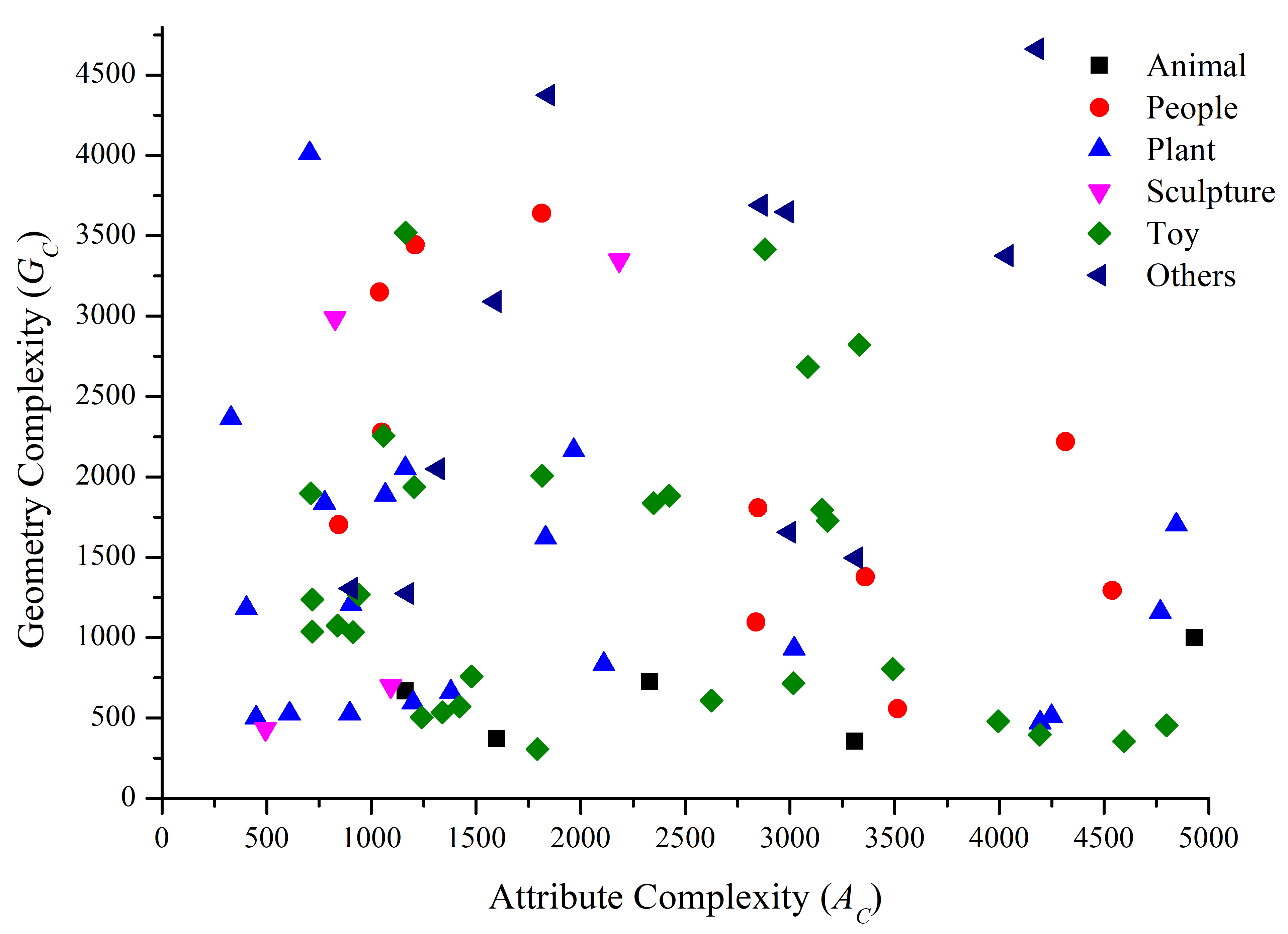}
		\end{minipage}
\label{fig12}
	}
	\centering
	\caption{Properties of point clouds in the proposed SYSU-PCUD. (a) Number of points, (b) Geometry and attribute complexities.}
	\label{Database}
\end{figure}

The collected point clouds are colorful and diverse in objects. Fig. \ref{fig4} illustrates some example point clouds in the dataset, which has 121 point clouds in six general categories, including animal, people, plant, sculpture, toy and others. We observe that these point clouds differ from each other in objects and contents. Meanwhile, they are colored, dense and of high quality. In addition, the scale of each point cloud is investigated. Fig. \ref{fig7} shows the number of points for each source point cloud in the SYSU-PCUD dataset. Most point clouds have 80,000 and 1,500,000 points. The largest one is the $54^{th}$ point cloud called \emph{Leap}, which has 3,817,422 points. The smallest one is the $39^{th}$ point cloud called \emph{Duck}, which has 19,247 points. The collected point clouds are colorful and diverse in the number of points, which are much larger than those in the existing up-sampling datasets \cite{[4]}\cite{[5]}.

To further illustrate the content diversity of the dataset, we analyzed the geometry complexity $G_{C}$ and the attribute complexity $A_{C}$ of each point cloud. We first projected the point cloud patches on the six faces of its bounding box \cite{[26]}, which generated six pairs of geometry and attribute images. Then, we compute the average $G_{C}$ and $A_{C}$ as
\begin{equation}
\begin{cases}
G_C = \frac{1}{M_iN}\sum\limits_{i=1}^{N} \sum\limits_{j=1}^{M_i} (g_{i,j} -\frac{1}{M_i} \sum\limits_{j=1}^{M_i}g_{i,j})^2\\
A_C = \frac{1}{M_iN}\sum\limits_{i=1}^{N} \sum \limits_{j=1}^{M_i} (a_{i,j} -\frac{1}{M_i} \sum\limits_{j=1}^{M_i}a_{i,j})^2
\end{cases},
\end{equation}
where $N$ is the number of projected images, which is 6; $g_{i,j}$ is the depth value corresponding to pixel $j$ in the $i$th projected geometry image, $a_{i,j}$ is the luminance value corresponding to pixel $j$ in the $i$-th projected attribute image, and $M_i$ is the number of pixels in the $i$-th image. Fig. \ref{fig12} shows the geometry and the attribute complexities of each point clouds, where horizontal and vertical axes are $A_{C}$ and $G_{C}$, respectively, and each point represents a point cloud. We can observe that the collected point clouds spread in a wide range, which indicates they are diverse in both geometry and attribute complexities. Overall, the point clouds in the database are not only diverse in categories and contents, but also diverse in the number of points, attribute and geometry complexities.

Then, all the collected source point clouds were down-sampled with rates 4$\times$, 8$\times$, 12$\times$, and 16$\times$, respectively. The source point clouds and their down-sampled versions compose the large scale point cloud dataset, which can be used to train, validate and test the geometry, the attribute and the joint up-sampling algorithms for point clouds. {The dataset is released to publicity \cite{SYSU-PCUD}.}

\begin{figure*}[h]
    \centering
    \includegraphics[width=0.75\textwidth]{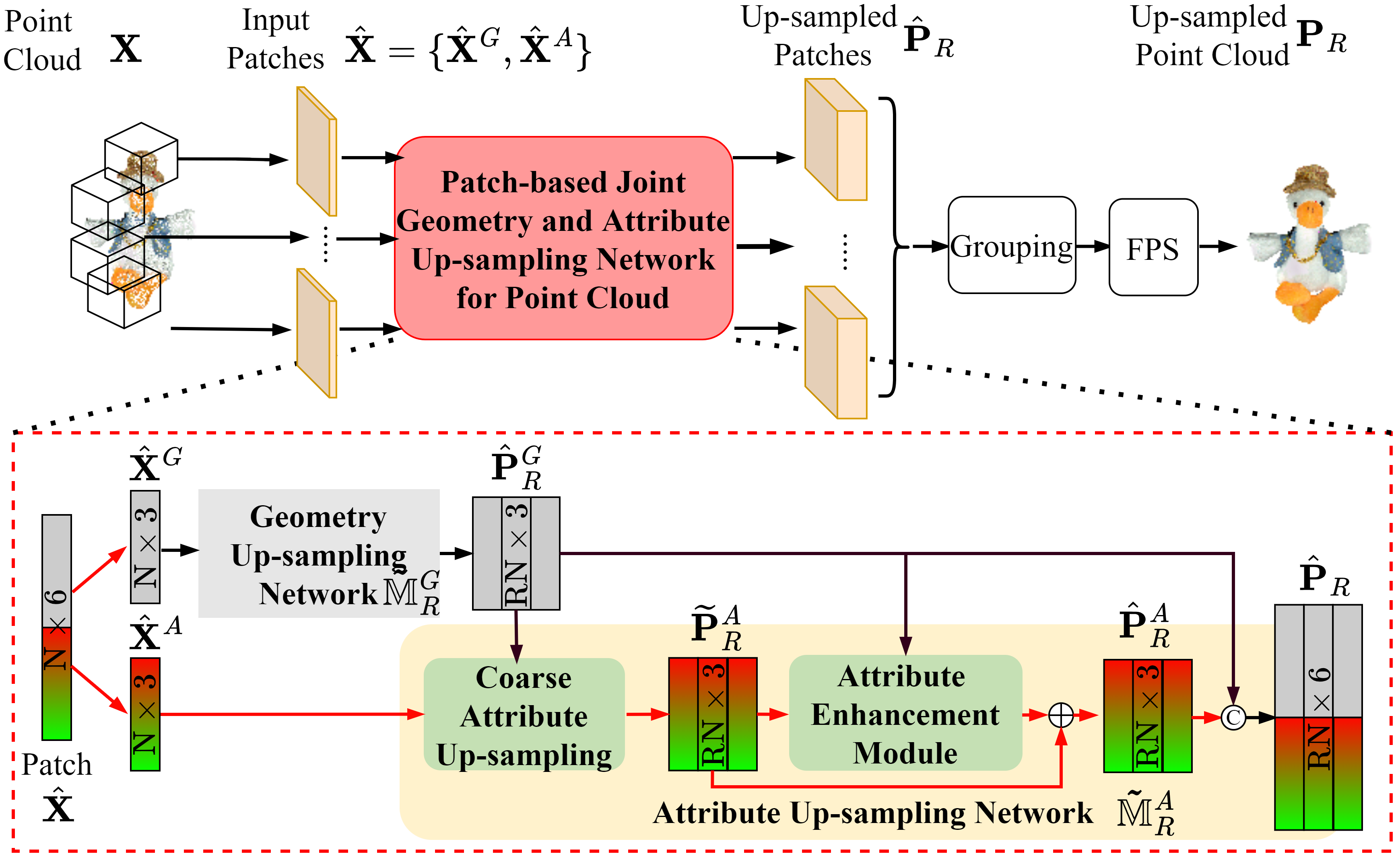}
    \caption{The proposed JGAU framework for large-scale colored point clouds.
    }
    \label{fig1}
\end{figure*}
\section{Problem Formulation for Point Cloud Up-sampling}
Unlike the conventional sparse colorless point clouds, large-scale colored point clouds are denser with more complicated structures and details, which is challenging to be up-sampled.
Let 
${\mathbf{X}= \{\mathbf{x}_i | \mathbf{x}_i \in \mathbb{R}^6, i=1,...,N\}}$
be a large-scale sparse colored point cloud with $N$ points. Each point $\mathbf{x}_i$ with index $i$ is a six dimensional vector consists of three dimensional geometry $\mathbf{x}_{i}^{G}\in \mathbb{R}^3$ and three dimensional attribute $\mathbf{x}_{i}^A\in \mathbb{R}^3$, i.e., $\mathbf{x}_i=\{\mathbf{x}_{i}^{G},\mathbf{x}_{i}^A\}$. We aim to generate a large-scale dense colored point cloud 
$ \mathbf{P}_R = \{\mathbf{p}_i^r | \mathbf{p}_i^r \in \mathbb{R}^6, i=1, ...,N, r=1, ...,R\}$
from the sparse $\mathbf{X}$ with up-sampling model $\mathbb{M}_R$,  
where the up-sampling factor is $R$. $\{\mathbf{p}_i^r\}$ is a set of $r$ times up-sampled points based on $\{\mathbf{x}_i\}$,
which contains more geometry and attribute details. The point cloud up-sampling problem is formulated as
\begin{equation}
\begin{cases}
\mathbf{P}_R=\mathbb{M}_R(\mathbf{X}) \\
\mathbb{M}_R^*=\arg\min\limits_{\mathbb{M}_R}\mathbb{D}(\mathbf{P}_R,\mathbf{P}_O)
\end{cases},
\label{eq_problem}
\end{equation}
where $\mathbb{D}$ is a high dimensional distance metric to measure the difference between the original and up-sampled point clouds $\mathbf{P}_O$ and $\mathbf{P}_R$. Eq. \eqref{eq_problem} means to find an optimal up-sampling model $\mathbb{M}_R$ to minimize the difference between $\mathbf{P}_O$ and $\mathbf{P}_R$.

The problem in Eq. \eqref{eq_problem} can be divided into two sub-problems, which are colored point cloud up-sampling and large scale up-sampling. Firstly, colored point clouds are six dimensional data, which consist of three dimensional geometry and three dimensional attribute, respectively, i.e., $\mathbf{X}=\{\mathbf{X}^G,\mathbf{X}^A\}$, $\mathbf{P}_O=\{\mathbf{P}^G_O,\mathbf{P}^A_O\}$, and $\mathbf{P}_R=\{\mathbf{P}_R^G,\mathbf{P}_R^A\}$. However, the property of geometry is different from that of attribute. Thus, the up-sampling for six dimensional point cloud $\mathbf{P}_R$ can be modelled as two sub-up-sampling problems of three dimensional data, which are
\vspace{0.5ex}
\begin{equation}
\begin{cases}
\mathbb{M}^{G*}_R=\arg\min\limits_{\mathbb{M}^G_R}\mathbb{D}^G(\mathbb{M}^G_R(\mathbf{X^G}),\mathbf{P}^G_O)\\
\mathbb{M}^{A*}_R=\arg\min\limits_{\mathbb{M}^A_R}\mathbb{D}^A(\mathbb{M}^A_R(\mathbb{M}^G_R(\mathbf{X^G}),\mathbf{X^A}),\mathbf{P}^A_O)
\end{cases},
\label{eq_sub_problem}
\end{equation}
where $\mathbb{M}^{\phi}_R$ is an up-sampling model at up-sampling rate $R$ for $\phi$ channel, $\mathbb{D}^{\phi}$ is a distortion metric that measures the difference between original point clouds $\mathbf{P}^{\phi}_O$ and up-sampled point clouds $\mathbf{P}^{\phi}_R$, $\mathbf{P}^{\phi}_R=\mathbb{M}^{\phi}_R(\mathbf{X^{\phi}})$, $\phi\in\{G, A\}$, $A$ and $G$ represent attribute and geometry. Eq. \eqref{eq_sub_problem} aims to find the optimal geometry up-sampling model $\mathbb{M}^G_R$ with sparse geometry $\mathbf{X^G}$ and the optimal attribute up-sampling model $\mathbb{M}^A_R$  based on up-sampled geometry $\mathbb{M}^G_R(\mathbf{X^G})$ and sparse attribute $\mathbf{X^A}$.

Secondly, with the development of deep learning technology, it can be used to learn {deep up-sampling models} for geometry and attribute, i.e., $\mathbb{M}^G_R$ and $\mathbb{M}^A_R$. However, due to the limited memory capacities of the GPU, the large-scale colored point cloud cannot be input and processed entirely. The large-scale point cloud can be divided into a number of small patches and processed one-by-one with a patch-based up-sampling, which reduces the memory consumption and enables parallel computation. Then, up-sampled patches are re-grouped to reconstruct the large-scale point clouds. This process is so-called ``divide-and-conquer'' strategy, which makes sense since the local neighborhood information will be mainly exploited in the up-sampling. Let $\hat{\mathbf{X}}$ be a smaller patch that generates the dense patch $\hat{\mathbf{P}}_{R}$ after up-sampling. Patches are six dimensional data, which consist of three dimensional geometry and three dimensional attribute, respectively, i.e., $\hat{\mathbf{X}}=\{\hat{\mathbf{X}}^G,\hat{\mathbf{X}}^A\}$
and $\hat{\mathbf{P}}_R=\{\hat{\mathbf{P}}^G_R,\hat{\mathbf{P}}^A_R\}$.
In this case, the up-sampling for a sub-set of points can be modelled as
\begin{equation}
\begin{cases}
\mathbb{\tilde{M}}^{G*}_{R}=\arg\min\limits_{\mathbb{\tilde{M}}^G_R}\mathbb{D}^G (\mathbb{G}^G(\mathbb{\tilde{M}}^G_R(\hat{\mathbf{X}}^G)),\mathbf{P}^G_O),\\
\mathbb{\tilde{M}}^{A*}_{R}=\arg\min\limits_{\mathbb{\tilde{M}}^A_R}\mathbb{D}^A(\mathbb{G}^A(\mathbb{\tilde{M}}^A_R(\mathbb{\tilde{M}}^G_R(\hat{\mathbf{X}}^G),\hat{\mathbf{X}}^A)),\mathbf{P}^A_O),
\end{cases}
\label{eq_sub_problem_patch}
\end{equation}
where $\mathbb{\tilde{M}}^{\phi}_{R}$ is the learned deep up-sampling model, $\phi\in\{G, A\}$, $\mathbb{G}()$ is a group operation that aggregates the up-sampled dense patches. Finally, the large-scale up-sampled dense and colored point cloud $\mathbf{P}_R$ is output as
\begin{equation}
\begin{cases}
{\mathbf{P}}_R= \{\mathbb{G}^G(\hat{\mathbf{P}}^G_R), \mathbb{G}^A(\hat{\mathbf{P}}^A_R)\} \\
\hat{\mathbf{P}}^A_R=\mathbb{\tilde{M}}^A_R(\hat{\mathbf{P}}^G_R,\hat{\mathbf{X}}^A)\\
\hat{\mathbf{P}}^G_R=\mathbb{\tilde{M}}^G_R(\hat{\mathbf{X}}^G)
\end{cases},
\label{Output}
\end{equation}
where $\hat{\mathbf{P}}^A_R$ and $\hat{\mathbf{P}}^G_R$ are up-sampled attribute and geometry patches through learned models $\mathbb{\tilde{M}}^A_R$ and $\mathbb{\tilde{M}}^G_R$, respectively. $R$ is the up-sampling rate. These patches are grouped and re-sampled to output $\mathbf{P}_R$.
However, one of the drawbacks of this ``divided and conquer'' strategy is the boundary regions cannot use effective information from neighboring patches, which will be solved by overlapped patch sampling. The key is to develop geometry and attribute up-sampling models, i.e., $\mathbb{\tilde{M}}^A_R$ and $\mathbb{\tilde{M}}^G_R$.

\section{Proposed JGAU for Colored Point Clouds}
\subsection{Framework of the Proposed JGAU}

The large-scale colored point clouds shall be divided into smaller patches, up-sampled one-by-one and then re-grouped to reconstruct the large-scale point clouds.
Fig. \ref{fig1} illustrates the framework and network architecture of the proposed JGAU, which consists of point cloud partitioning, patch-based JGAU network, regrouping and sampling. The large-scale point cloud $\mathbf{X}$ is divided into smaller patches by the ``divide and conquer" based partitioning.
Each patch $\hat{\mathbf{X}}$ is input into patch-based point cloud up-sampling network to generate a dense patch with attribute details.
Patch-based JGAU network mainly consists of a geometry up-sampling network $\mathbb{\tilde{M}}^G_R$ and an attribute up-sampling network $\mathbb{\tilde{M}}^A_R$, which includes a coarse attribute up-sampling and an AEM. Finally, the up-sampled point cloud patches are re-grouped and sampled to a desired ratio with the FPS to generate large-scale dense point cloud $\mathbf{P}_R$.

In the patch-based JGAU network, the geometry up-sampling generates dense $\hat{\mathbf{P}}_R^G$ from sparse $\hat{\mathbf{X}}^G$. Due to the superior performance of the Dis-PU \cite{[24]} in geometry up-sampling, it is applied to $\mathbb{\tilde{M}}^G_R$ and up-sample the geometry component of partitioned point cloud patches. Since the attribute is attached to the geometry, the attribute up-sampling $\mathbb{\tilde{M}}^A_R$ is guided by the up-sampled geometry. In the proposed $\mathbb{\tilde{M}}^A_R$, based on $\hat{\mathbf{P}}_R^G$, the sparse attribute $\hat{\mathbf{X}}^A$ is coarsely up-sampled to generate $\widetilde{\mathbf{P}}_R^A$ by the coarse attribute up-sampling. By further exploiting high-dimensional local attribute features, the AEM is proposed to refine the coarse attribute $\widetilde{\mathbf{P}}_R^A$. Then, attribute offsets are obtained by regressing attribute features, which generates a fine-grained attribute set $\hat{\mathbf{P}}_R^A$. Finally, $\hat{\mathbf{P}}_R^G$ and $\hat{\mathbf{P}}_R^A$ are concatenated to generate dense patch $\hat{\mathbf{P}}_R$. Details of the key modules in the patch-based JGAU are presented in the following subsections.

\subsection{Point Cloud Partitioning and Overlap Ratio}
\label{PCpartion}
Based on Eq. \eqref{eq_sub_problem_patch}, we need to divide the large-scale point clouds into smaller patches to train the up-sampling network and do the up-sampling. However, although partitioning the point cloud into smaller patches may reduce the memory and computations required in the network training, it may decrease the quality of the up-sampled point cloud in the boundary regions of the patches due to insufficient textural information. To have a good trade-off between the memory capacity of training and the quality of up-sampling, it is important to determine the number of patches ($N$). In addition, to improve the up-sampling quality at the boundary regions, the point cloud is divided into overlapped patches with a ratio $c$, representing the overlapping ratio of points among patches. Thus, the number of patches $N$ is
\begin{equation}
N=\frac{n \times c}{m},
\label{patch}
\end{equation}
where $n$ represents the total number of points in the point cloud, and $m$ is the number of points in each patch.
Increasing batch size $m$ to get a larger patch will take up more GPU memory capacity and increase the available local neighborhood information for up-sampling. To have a good trade-off between the two aspects, $m$ is set as 256. Increasing the overlap ratio $c$ may improve the up-sampling quality, but may cause additional computational complexity as more points and patches shall be processed. $c$ is set as 300\% to have a good trade-off between computational complexity and quality. The largest point cloud contains 3,817,422 points and the smallest point cloud contains 19,247 points. So, $N$ varies from 256 to 44736 depending on the number of points $n$ and ratio $c$.

\begin{figure}[t]
\centering
\subfigure[Statue (29.85 dB)]{
	\begin{minipage}[t]{0.3\linewidth}
		\centering
		\includegraphics[trim= 50 0 0 0,scale=0.15]{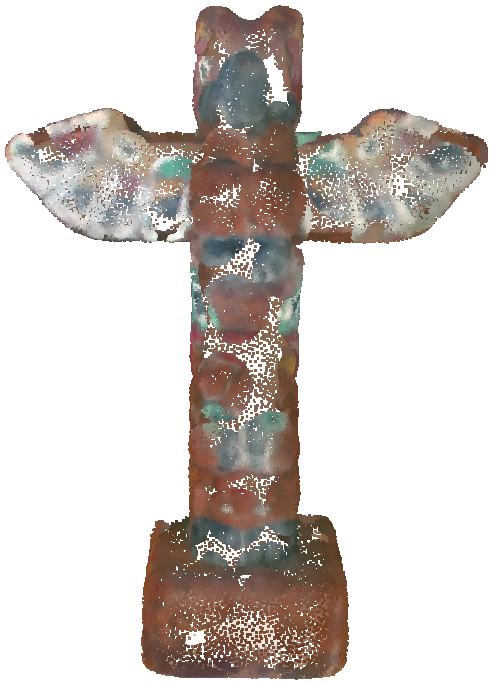}
	\end{minipage}
}%
\subfigure[Duck (20.68 dB)]{
	\begin{minipage} [t]{0.3\linewidth}
		\centering
		\includegraphics[trim= 40 0 0 0,scale=0.15]{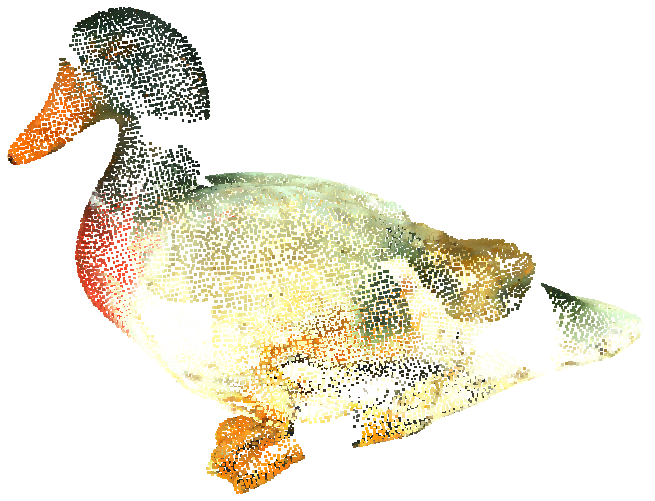}
	\end{minipage}%
}%
\subfigure[Doll8 (31.95 dB)]{
	\begin{minipage}[t]{0.3\linewidth}
		\centering
		\includegraphics[trim= 0 0 0 0,scale=0.15]{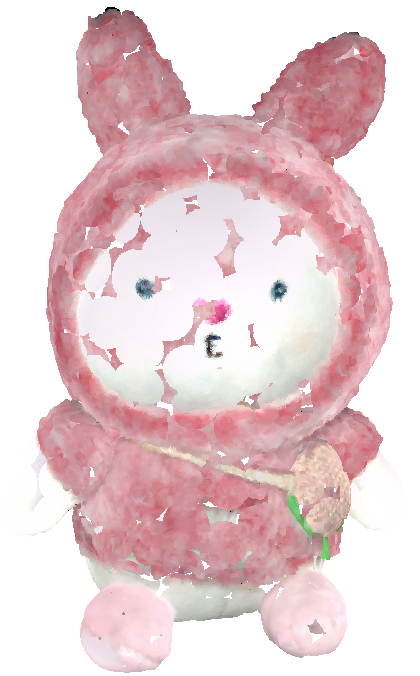}
	\end{minipage}
}%
\\
\subfigure[Statue (34.00 dB)]{
	\begin{minipage}[t]{0.3\linewidth}
		\centering
		\includegraphics[trim= 50 0 0 0,scale=0.15]{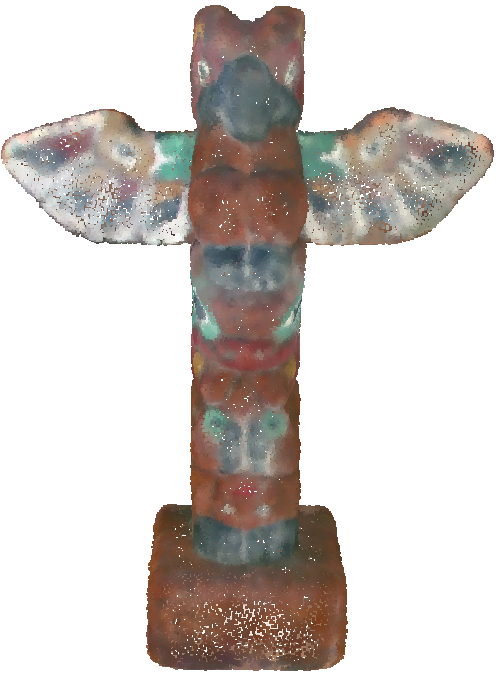}
	\end{minipage}
}%
\subfigure[Duck (21.51 dB)]{
\begin{minipage} [t]{0.3\linewidth}
\centering
\includegraphics[trim= 40 0 0 0,scale=0.15]{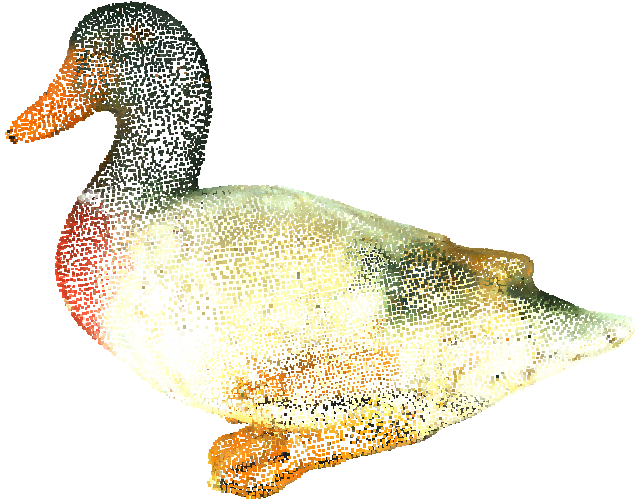}
\end{minipage}%
}%
\subfigure[Doll8 (33.51 dB)]{
	\begin{minipage}[t]{0.3\linewidth}
		\centering
		\includegraphics[trim= 0 0 0 0,scale=0.15]{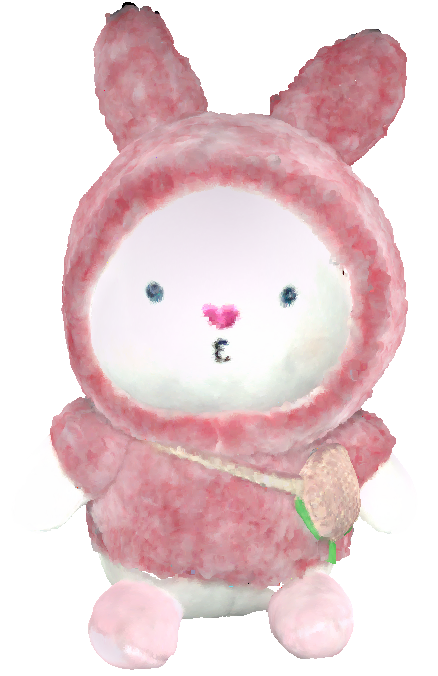}
	\end{minipage}
}%
\centering
\caption{Results of up-sampled point cloud with/without overlapped patches. (a)-(c) without overlapped patches, (d)-(f) with overlapped patches.}
\label{overlap}
\end{figure}

To validate the effectiveness of the point cloud partitioning and overlapping method, we performed an up-sampling experiment, where two schemes, up-sampling with and without overlapped patches, were evaluated. Three point clouds, $Statue$,  $Duck$ and $Doll8$ were up-sampled.
Fig. \ref{overlap} shows the visualization results of up-sampled point cloud using overlapped and non-overlapped patches. If without using the overlapping, attribute Peak Signal-to-Noise Ratio (PSNR) of the up-sampled $Statue$ is 29.85 dB and there are noticeable holes and artifacts at the patch boundaries. On the other hand, while using the overlapping, the attribute PSNR of the up-sampled $Statue$ is 34.00 dB, which is 4.15 dB higher than that of without overlapping. Similarly, by using the proposed overlapped partitioning method, significant PSNR gains and visual quality improvement can be achieved for $Duck$ and $Doll8$. These results demonstrate that the overlapped partitioning is effective for large-scale point cloud processing.

\subsection{Coarse Attribute Up-sampling}
\begin{figure}[t]
	\centering
	\subfigure[]{
		\begin{minipage}[t]{1.0\linewidth}
			\centering
			\includegraphics[width=1.0\textwidth]{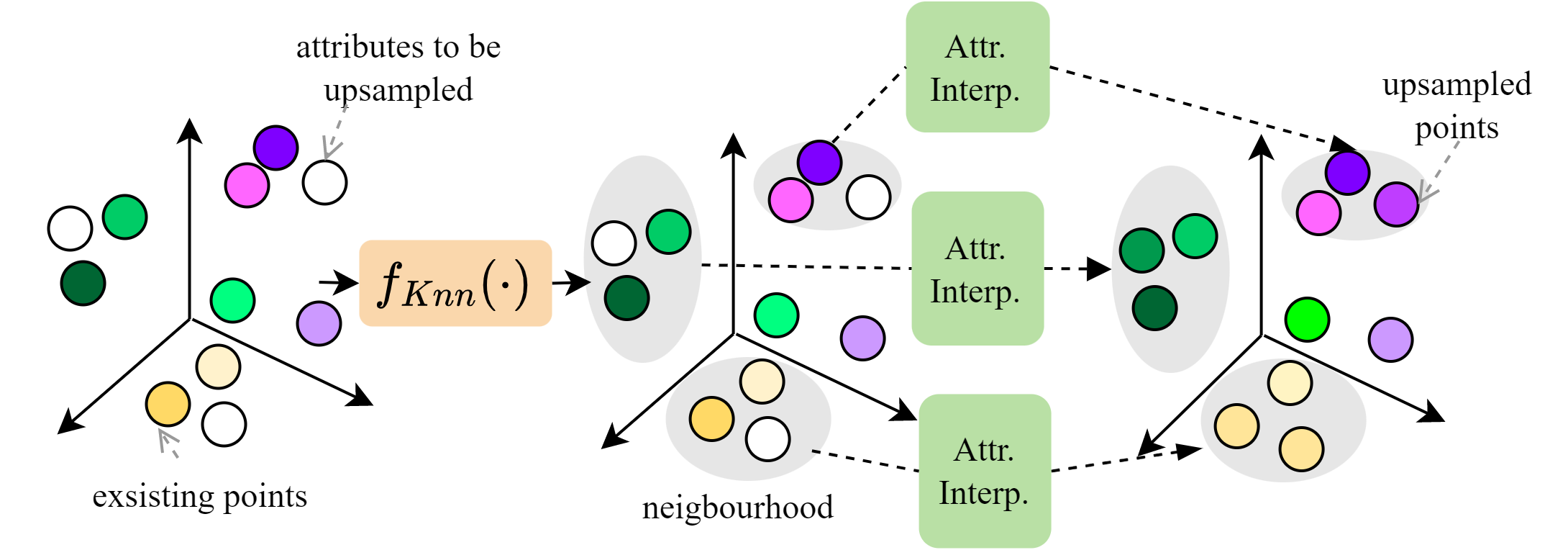}
		\end{minipage}
	}
	\subfigure[]{
		\begin{minipage}[t]{1.0\linewidth}
			\centering
			\includegraphics[width=0.95\textwidth]{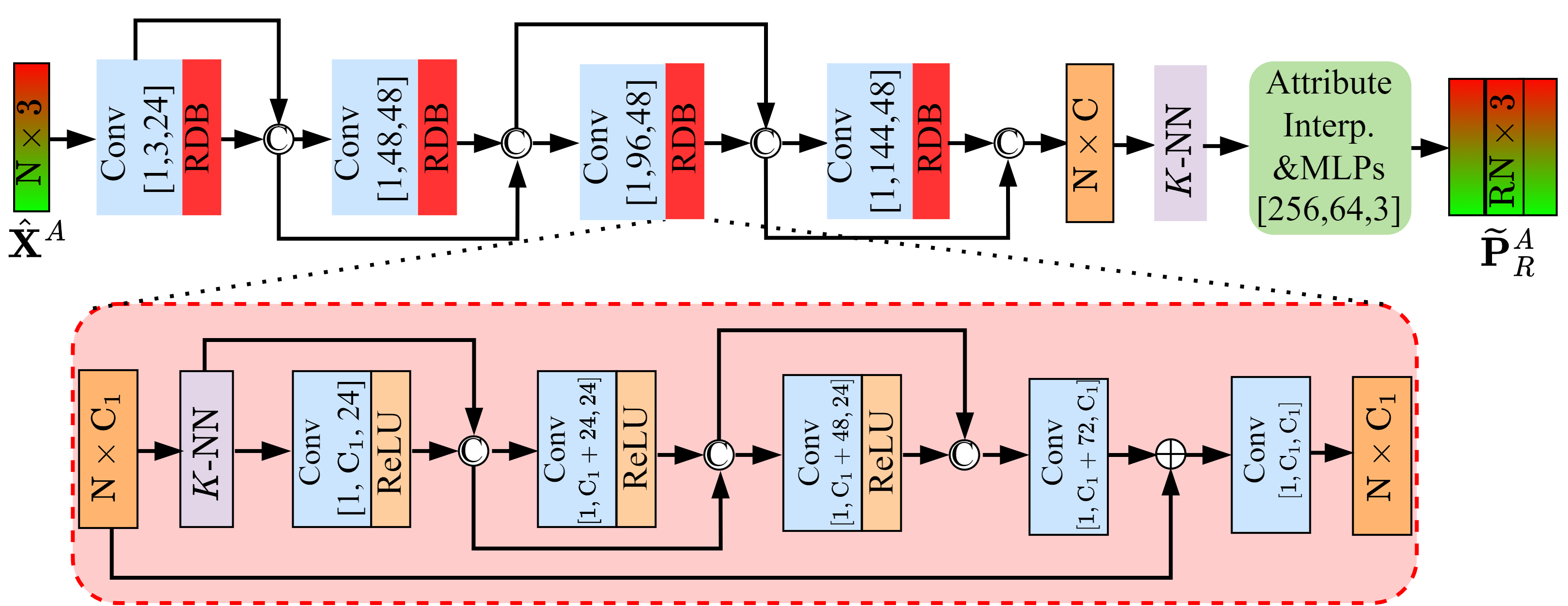}
		\end{minipage}
	}
	\centering
	\caption{Coarse attribute up-sampling. (a) Flowchart, (b) architecture of DLAI.}
	\label{Attribute_intr}
\end{figure}

To up-sample the attribute attached to each point, we propose a coarse attribute up-sampling on the basis of the geometry information. Fig.\ref{Attribute_intr} (a) shows the flowchart of the coarse attribute up-sampling. Firstly, dense up-sampled geometry is grouped with $K$-NN $f_{KNN}(\cdot)$, and then empty attribute points are interpolated based on the geometrical and attribute information in each group. We propose two sub-algorithms for the attribute interpolation, which are Geometric Distance Weighted Attribute Interpolation (GDWAI) and Deep Learning based Attribute Interpolation (DLAI). Finally, the coarsely up-sampled attribute is refined by AEM.
\begin{figure*}[t]
    \centering
    \includegraphics[width=0.95\textwidth]{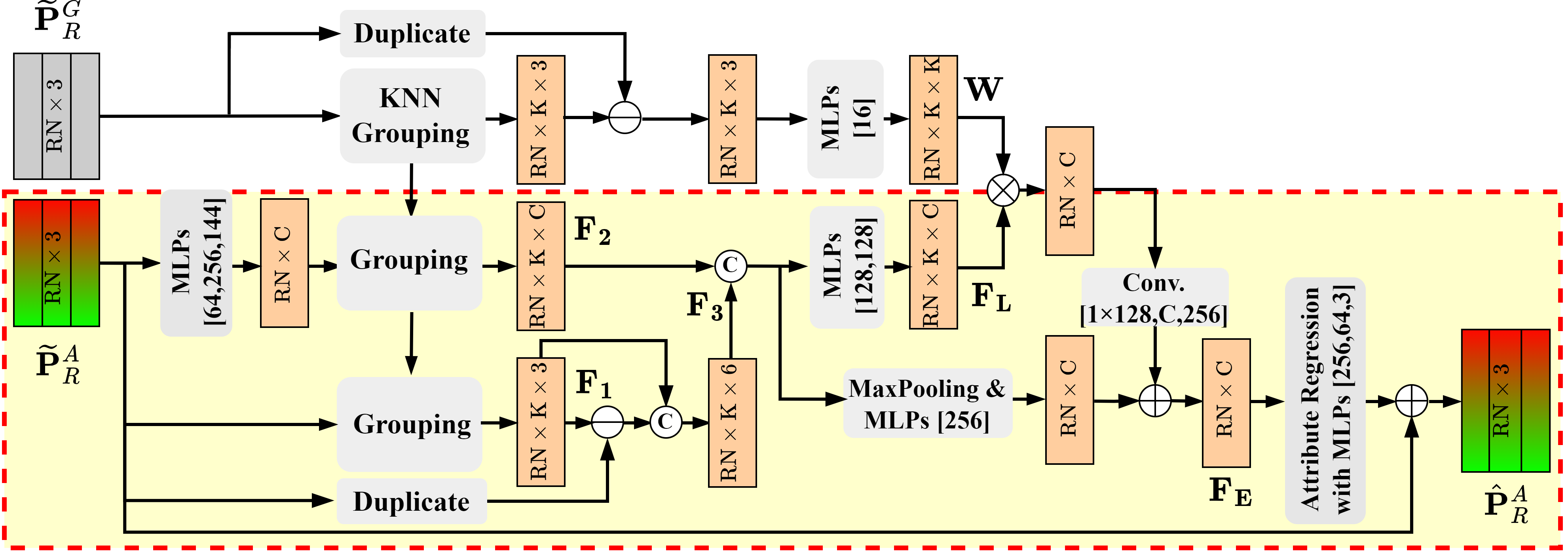}
    \caption{Architecture of the AEM using auxiliary geometry information, where the yellow shadow part is specified for the attribute enhancement, and dark yellow rectangles specify the dimensions of feature.}
    \label{fig3}
\end{figure*}
\subsubsection{GDWAI}
According to the attribute correlation of points in the $K$-nearest neighborhood in the point cloud, we propose the GDWAI to generate the coarse attribute with auxiliary geometry information. $K$-NN is used to calculate the $k_1$ nearest neighboring points and extract their attribute of each point. Due to the spatial correlation, the attribute is highly correlated among neighboring points. This spatial correlation generally decreases as the geometrical distance goes far. Therefore, the geometric distances of the $k_1$ points are then used as the importance weights in the attribute interpolation. {The attribute of the $j$-th point to be interpolated ${\mathbf{x}}_{j}^A$ is calculated as
 \begin{equation}
{\mathbf{x}}_{j}^A=\frac{1}{\sum_{i=1}^{k_1} w_i}\sum_{i=1}^{k_1} (w_i\mathbf{x}^A_i),
\label{pcolor}
\end{equation}
where $w_i$ is a geometric distance based importance weight, which is a reciprocal to the point-to-point Euclidean distance \cite{[14]} between the point to be interpolated and its $i$-th neighboring point. $\mathbf{x}^A_i$ is the attribute of point $i$ in $\hat{\mathbf{X}}^A$, $k_1$ is the number of neighboring points for $j$-th point via KNN. Then, the empty attribute of the dense point clouds can be filled one-by-one by using Eq.\eqref{pcolor}, which is presented as $\widetilde{\mathbf{P}}_R^A-\hat{\mathbf{X}}^A=\{{\mathbf{x}}_{j}^A\}$.}


\subsubsection{DLAI}

In addition to distance importance, the spatial correlation of point cloud attribute also varies with object contents, which reduces the object boundaries and textural regions. Therefore, in addition to the GDWAI, we propose a DLAI to exploit the attribute pattern in the point cloud while generating the coarse attribute set. Fig. \ref{Attribute_intr} (b) shows the network structure of the proposed DLAI, { where the K-NN and attribute interpolation modules are supported by the geometry information $\hat{\mathbf{X}}_G$. The numbers in the parentheses of convolution represent kernel size, the numbers of input and output channels, while the numbers of the MLP module represent the output channels of each layer. We introduce convolutional networks and the Residual Dense Block (RDB) \cite{[11]} to extract high-dimensional attribute features from the local neighborhood obtained by $K$-NN.
Specifically, we use four layers of CNN and RDB networks with kernel size 1$\times$1 to extract high-dimensional inartistic and neighborhood attribute features. $C_1$ in RDB is the dimensions of data that depends on the output feature of previous convolution layer. And then a $K$-NN is used to aggregate the $k_1$ locally neighborhood points for attribute interpolation, where geometric distances are used as the importance weights of the high-dimensional attribute features. To further exploit the spatial correlation, KNN is performed in each RDB sub-network. Finally, MLPs are used to regress the high-dimensional attribute features and up-sample dense attribute.}

To determine the optimal $k_1$ value for the $K$-NN, attribute up-sampling experiments using the GDWAI and DLAI were performed. We set the candidate values of $k_1$ as 1, 2, 3, and 4, respectively. We randomly selected three point clouds for 4$\times$ attribute up-sampling.
Table \ref{attrbute_up} shows the comparison results of interpolation experiments with different $k_1$ values, where the values in bold denote the best.
Based on the up-sampled point clouds, the average PSNR achieved by GDWAI were 33.03 dB, 33.07 dB, and 32.75 dB when $k_1$ is 1, 3, and 4, respectively, and 33.53 dB when $k_1$ is 2, which was 0.50 dB, 0.46 dB, and 0.78 dB higher when $k_1$ is 1, 3, and 4, respectively.
For the DLAI, the PSNR values were 33.80 dB, 34.01 dB, 33.54 dB, and 33.24 dB when $k_1$ changes from 1 to 4, respectively. When $k_1$ is 2, the PSNR of up-sampled point clouds achieved by the DLAI was 0.48 dB higher than that of the GDWAI. Overall, it is found that the GDWAI and DLAI achieve their best when $k_1$ is 2. Moreover, the DLAI is superior to GDWAI by additionally learning attribute patterns.

\begin{table}[t]
	\centering
	\caption{PSNRs of GDWAI and DLAI using different $K$ of $K$-NN.}
	\resizebox{\linewidth}{!}{
		\begin{tabular}{|c|c|c|c|c|c|}
			\hline
			Point Clouds & Methods & $k_1$ = 1    & $k_1$ = 2    & $k_1$ = 3    & $k_1$ = 4\\
			\hline
			$Banana$ & GDWAI & 33.62  & \textbf{34.10} & 33.53  & 33.12\\
            \hline
			$Doll$  & GDWAI & 33.52  & \textbf{34.10} & 33.79  & 33.56\\
            \hline
			$Longdress$ & GDWAI & 31.96  & \textbf{32.39} & 31.90  & 31.58\\
			\hline
			\multicolumn{2}{|c|}{Average}  & 33.03  & \textbf{33.53} & 33.07  & 32.75\\
			\hline
			$Banana$ & DLAI  & 34.02  & \textbf{34.19} & 33.81  & 33.12\\
            \hline
			$Doll$  & DLAI  & 34.31  & \textbf{34.53} & 34.19  & 34.06\\
            \hline
			$Longdress$ & DLAI  & 33.08  & \textbf{33.30} & 32.61  & 32.55\\
			\hline
			\multicolumn{2}{|c|}{Average}  & 33.80  & \textbf{34.01} & 33.54  & 33.24\\
			\hline
		\end{tabular}
	}
	\label{attrbute_up}
\end{table}

\subsection{AEM and Attribute Regression}

After the coarse attribute up-sampling, some attribute details are still missing in the up-sampled point cloud. To handle this problem, we proposed the AEM to enhance the coarsely up-sampled attribute. Fig. \ref{fig3} shows the flowchart of the proposed AEM, which exploits the neighborhood correlation of attribute to enhance $\widetilde{\mathbf{P}}_R^A$ with auxiliary geometry information. Similar to the coarse attribute up-sampling, both the geometrical distance and attribute patterns are exploited in the AEM while enhancing the attribute quality.

Firstly, the geometry distance of a point is used as an importance weight for attribute refinement. The geometrical importance weight $\mathbf{W}$ is generated from the difference between the duplicated $\hat{\mathbf{P}}^G_R$ and its $K$-NN neighborhood by
\begin{equation}
{
\mathbf{W}= f_{M}(\hat{\mathbf{P}}^G_R-f_{KNN}(\hat{\mathbf{P}}^G_R)),
}
\end{equation}
where $f_{M}()$ is the MLPs and $f_{KNN}()$ is the $K$-NN clustering. The optimal parameter of the $K$-NN clustering $f_{KNN}()$, $k_2$, will be experimentally analyzed in Section V.F. Distant points usually has lower importance in the attribute enhancement.

Secondly, in the attribute refinement, high-dimensional attribute features are extracted from $\widetilde{\mathbf{P}}_R^A$ based on $K$-NN and MLPs for feature extraction, which is $\mathbf{F}_2 = f_{KNN}(f_{M}(\widetilde{\mathbf{P}}^A_R))$. Since the attribute is attached to the geometry, the extracted attribute features are grouped with $K$-NN under the guidance of geometry. Then, the local attribute information $\mathbf{F}_{1}$ is obtained by $K$-NN grouping of the attribute $\widetilde{\mathbf{P}}^A_R$. Meanwhile, $\widetilde{\mathbf{P}}_R^A$ is duplicated $K$ times, subtracted by $\mathbf{F}_{1}$, and then concatenated with $\mathbf{F}_{1}$ to obtain the local neighboring features $\mathbf{F}_{3}$. Then, the high-level attribute features $\mathbf{F}_L \in \mathbb{R}^{R N \times K \times C}$ is obtained from concatenated local features $\mathbf{F}_3$ and high dimensional pattern features $\mathbf{F}_2$ using MLPs, which are
\begin{equation}
{
\begin{cases}
\mathbf{F}_L=f_{M}(f_{C}(\mathbf{F}_2,\mathbf{F}_3))\\
\mathbf{F}_3=f_{C}(f_{KNN}(\widetilde{\mathbf{P}}^A_R)-\widetilde{\mathbf{P}}^A_R,f_{KNN}(\widetilde{\mathbf{P}}^A_R))\\
\end{cases},
}
\end{equation}
where $f_{C}()$ is a concatenate operation.

Finally, the distance weight $\mathbf{W}$ is multiplied by the attribute feature $\mathbf{F}_{L}$ to generate locally important features. Residual learning is employed to obtain attribute offsets instead of enhanced attribute values. The reason is that the absolute attribute value distribution is more diverse and has a wider distribution when compared with the relative offset. Attribute offsets are generated using attribute regression i.e., MLPs, for $\mathbf{F}_E$. The coarse attribute set $\widetilde{\mathbf{P}}_R^A$ and the attribute offsets are added to obtain the fine-grained attribute set $\hat{\mathbf{P}}_R^A$ as
\begin{equation}
{
\begin{cases}
\hat{\mathbf{P}}^A_R=f_{M}(\mathbf{F_E})+\widetilde{\mathbf{P}}^A_R\\
\mathbf{F_E}=f_{MP}(f_{C}(\mathbf{F}_2,\mathbf{F}_3))+f_{Conv}(\mathbf{W}\times\mathbf{F}_L)\\
\end{cases},
}
\end{equation}
where $f_{MP}()$ represents MLPs and maxpooling, $f_{Conv}()$ is a convolution operation.

\subsection{Training Strategy and Loss Function}

In the JGAU, the point cloud geometry up-sampling network and the attribute up-sampling network are trained jointly in an end-to-end manner to maximize the performance. Based on Eq. \eqref{eq_sub_problem}, the objectives of learning the geometry and attribute up-sampling networks for the point cloud are different. Dual optimizers shall be used to optimize the network training. However, objectives in Eq. \eqref{eq_sub_problem} are point cloud-level optimizations. Although re-grouped patches to point cloud will improve the up-sampling accuracy for the large-scale point clouds, it consumes a much larger memory capacity, causes more computational complexity of each iteration and also reduces the number of available training data, making the training infeasible. To handle this problem, Eq.\eqref{eq_sub_problem} is modified into patch-level objectives as
\begin{equation}
\begin{cases}
\mathbb{\tilde{M}}^{G*}_{R}=\arg\min\limits_{\mathbb{\tilde{M}}^G_R}\mathbb{D}^G (\mathbb{\tilde{M}}^G_R(\hat{\mathbf{X}}^G),\hat{\mathbf{X}}^G_O),\\
\{\mathbb{\tilde{M}}^{A*}_{R},\mathbb{\tilde{M}}^{G*}_{R}\}=\arg\min\limits_{\{\mathbb{\tilde{M}}^{A}_{R},\mathbb{\tilde{M}}^{G}_{R}\}}\mathbb{D}^A(\mathbb{\tilde{M}}^A_R(\mathbb{\tilde{M}}^G_R(\hat{\mathbf{X}}^G),\hat{\mathbf{X}}^A),\hat{\mathbf{X}}^A_O),
\end{cases}
\label{eq_sub}
\end{equation}
where $\hat{\mathbf{X}}^A_O$ and $\hat{\mathbf{X}}^G_O$ are patches partitioned by Subsection \ref{PCpartion} from the ground truth point cloud ${\mathbf{P}}_O$=$\{\mathbf{P}^A_O, \mathbf{P}^G_O\}$, respectively. $\mathbb{D}^G$ is a distance measurement for geometry, which uses Chamfer Distance (CD) \cite{[24]}, and $\mathbb{D}^A$ is a difference measurement, which is set as Mean Absolute Error (MAE) for simplicity. $\{\mathbb{\tilde{M}}^{A*}_{R},\mathbb{\tilde{M}}^{G*}_{R}\}$ is the learned JGAU model and $R$ is the up-sampling rate. $\mathbb{\tilde{M}}^{G*}_{R}$ is firstly trained using the first equation in Eq. \eqref{eq_sub}, then, $\{\mathbb{\tilde{M}}^{A*}_{R},\mathbb{\tilde{M}}^{G*}_{R}\}$ are jointly trained or refined using the second equation in Eq. \eqref{eq_sub}.


%

\section{Experimental Results and Analysis}

To validate the performance of the proposed JGAU methods, experiments were performed to up-sample the point clouds and compared with the state-of-the-art up-sampling methods. Firstly, the experimental settings, including database, methods, and evaluation metrics, are presented. Then, the performances of the attribute up-sampling and geometry up-sampling are evaluated. Finally, cross-validations among different training and testing sets, computational complexity analyses, and ablation studies are performed to evaluate the key modules and robustness of the proposed JGAU methods.

\subsection{Experimental Settings}
\subsubsection{Training and Testing Database}
In Section \ref{Sec:database}, we have built and released a large-scale colored point cloud dataset for up-sampling SYSU-PCUD, which has 121 source point clouds with diverse contents, various numbers of points and four sampling rates, i.e., 4$\times$, 8$\times$, 12$\times$, and 16$\times$. In the database, 75 point clouds (about 60\%) were used in the model training and the rest 46 point clouds (40\%, denoted as Set1) were randomly selected in the testing. Note that the source point clouds in the testing set have never been used in the training set. Moreover, the \emph{Others} category contains many types with only one point cloud, which were used to testify the generalization capability of the models. For the testing dataset, we directly used the 46 point clouds and their four randomly down-sampled point clouds. In the cross-validation experiment, training and testing point clouds were shuffled.
{In addition, two additional point cloud databases were used as testing for cross database validation. 25 point clouds from JPEG Pleno dataset \cite{JPEG-Pleno} were used as testing data for cross-database validation, which validated the generalization capability. Moreover, to validate the robustness to distortion, four point clouds from 8iVFBv2 dataset \cite{8iVFB} were compressed by Video-based Point Cloud Compression (V-PCC)\cite{V-PCC,ZhangTIP2023} and then up-sampled by using the proposed JGAU and benchmark schemes.}

\begin{table*}[t]
	\centering
	\caption{PSNR comparisons of the up-sampled point cloud by MFU \cite{[6]}, FSMMR \cite{[38]}, FGTV \cite{[44]}, {CU-NET \cite{[46]}, IT-DL-PCC \cite{IT-DL-PCC}, JGAU-GDWAI and JGAU-DLAI in the proposed SYSU-PCUD dataset}, where the best is in bold.[Unit:dB]}
		\setlength{\tabcolsep}{0.2mm}
	\scalebox{0.85}{	\begin{tabular}{|c|c|c|c|c|c|c|c|}
		\hline
		{ {Schemes}}         & { {MFU}}                                                                                                     & { {FSMMR}}                                                                                                   & { {FGTV}}                                                                                                    & { {CU-NET}}                                                                                       & { {IT-DL-PCC}}                                                                                   & { {JGAU-GDWAI}}                                                                                              & { {JGAU-DLAI}}                                                                                                        \\ \hline
		{ {Point Clouds}}    & { 4$\times$/8$\times$/12$\times$/16$\times$}                                                                                                    & { 4$\times$/8$\times$/12$\times$/16$\times$}                                                                                                    & { 4$\times$/8$\times$/12$\times$/16$\times$}                                                                                                    & { 2$\times$/5$\times$/10$\times$}                                                                                            & { 4$\times$/8$\times$/16$\times$}                                                                                            & { 4$\times$/8$\times$/12$\times$/16$\times$}                                                                                                    & { 4$\times$/8$\times$/12$\times$/16$\times$}                                                                                                             \\ \hline
		{ {Axeguy}}          & { 31.17/29.35/28.50/27.36}                                                                                          & { 30.76/29.04/28.06/27.20}                                                                                          & { 26.94/24.64/23.58/22.95}                                                                                          & { 35.34/29.88/26.87}                                                                                    & { 29.28/25.91/25.10}                                                                                    & { 33.41/31.40/30.47/\textbf{29.50}}                                                                                          & { \textbf{33.50/31.62/30.52}/29.15}                                                                                          \\ \hline
		{ {Banana}}          & { 30.60/28.16/27.69/27.67}                                                                                          & { 30.29/29.03/27.94/28.19}                                                                                          & { 32.04/29.84/28.88/28.32}                                                                                          & { 35.27/30.04/28.24}                                                                                    & { 26.46/24.52/22.68}                                                                                   & { 34.10/\textbf{31.94/30.64/30.15}}                                                                                          & { \textbf{34.19}/31.76/30.58/29.62}                                                                                          \\ \hline
		{ {Basketball}}      & { 39.13/37.35/36.37/35.68}                                                                                          & { 18.46/17.89/17.68/17.65}                                                                                          & { 38.96/36.68/35.48/34.65}                                                                                          & { 41.83/36.66/33.38}                                                                                    & { -/-/-}                                                                                                & { 41.49/39.70/38.66/37.76}                                                                                          & { \textbf{42.09/40.16/39.03/38.23}}                                                                                          \\ \hline
		{ {Cake}}            & { 32.97/31.28/30.29/29.62}                                                                                          & { 32.34/30.29/28.87/29.16}                                                                                          & { 24.67/21.42/19.94/19.04}                                                                                          & { 38.34/32.49/29.07}                                                                                    & { 29.74/26.51/25.33}                                                                                    & { 35.01/33.38/32.49/31.83}                                                                                          & { \textbf{35.24/33.59/32.62/31.88}}                                                                                          \\ \hline
		{ {CoffeeCup}}       & { 31.43/29.22/27.25/26.76}                                                                                          & { 30.94/28.89/27.92/27.26}                                                                                          & { 22.67/19.58/18.32/17.54}                                                                                          & { 38.02/31.01/26.70}                                                                                    & { 27.38/22.73/21.39}                                                                                    & { 33.81/31.48/30.31/29.45}                                                                                          & { 34.52/\textbf{32.06/30.57/29.54}}                                                                                                   \\ \hline
		{ {Dancer}}          & { 38.62/36.82/35.78/35.08}                                                                                          & { 18.81/18.18/18.01/17.92}                                                                                          & { 38.71/36.38/34.97/34.12}                                                                                          & { 41.50/36.21/32.87}                                                                                    & { -/-/-}                                                                                                & { 41.03/39.16/37.95/37.12}                                                                                          & { \textbf{41.60/39.58/38.44/37.47}}                                                                                          \\ \hline
		{ {Do1l8}}           & { 31.18/29.66/29.08/28.50}                                                                                          & { 28.32/27.43/27.22/26.75}                                                                                          & { 28.33/25.88/24.66/23.97}                                                                                          & { 31.77/28.79/26.62}                                                                                    & { -/-/-}                                                                                                & { 33.26/31.51/30.86/30.24}                                                                                          & { \textbf{33.51/31.73/30.94/30.26}}                                                                                          \\ \hline
		{ {Exercise}}        & { 39.76/38.39/37.47/36.92}                                                                                          & { 19.76/19.12/18.97/18.90}                                                                                          & { 40.16/38.15/37.11/36.36}                                                                                          & { 42.64/38.27/35.42}                                                                                    & { -/-/-}                                                                                                & { 41.80/40.26/39.36/38.88}                                                                                          & { \textbf{42.20/40.50/39.54/39.01}}                                                                                          \\ \hline
		{ {GlassesCase}}     & { 27.45/25.48/24.46/23.77}                                                                                          & { 26.76/24.98/24.00/23.41}                                                                                          & { 21.15/18.42/17.31/16.71}                                                                                          & { 33.67/27.10/23.81}                                                                                    & { 25.03/21.09/20.20}                                                                                    & { 29.58/27.43/26.38/25.65}                                                                                          & { \textbf{29.95/27.71/26.60/25.79}}                                                                                          \\ \hline
		{ {HoneydewMelon}}  & { 31.42/30.27/29.62/29.33}                                                                                          & { 31.00/29.95/29.51/29.11}                                                                                          & { 31.14/29.67/29.06/28.72}                                                                                          & { 37.42/{32.67}/30.84}                                                                                    & { 29.72/25.51/23.28}                                                                                    & { 33.09/31.91/31.22/30.85}                                                                                          & { \textbf{33.23}/\textbf{31.92}/\textbf{31.26/30.87}}                                                                                          \\ \hline
		{ {Leap}}            & { 36.24/33.65/32.42/31.50}                                                                                          & { 36.41/33.86/32.52/31.68}                                                                                          & { 24.55/21.94/20.63/19.75}                                                                                          & { 50.35/{39.81}/34.52}                                                                                    & { 29.42/24.37/21.90}                                                                                    & { 38.14/35.92/34.52/33.46}                                                                                          & { \textbf{38.74}/\textbf{36.58}\textbf{/35.19/34.07}}                                                                                          \\ \hline
		{ {Longdress}}       & { 30.24/28.08/27.01/26.31}                                                                                          & { 29.51/27.36/26.47/25.81}                                                                                          & { 29.20/26.98/25.86/25.22}                                                                                          & { 33.48/27.78/25.30}                                                                                    & { 27.05/24.70/23.97}                                                                                    & { 32.39/29.78/28.55/27.71}                                                                                          & { \textbf{33.30/30.30/28.86/27.72}}                                                                                          \\ \hline
		{ {Mask}}            & { 33.12/31.12/30.79/30.09}                                                                                          & { 33.31/31.53/30.94/30.19}                                                                                          & { 33.41/31.69/30.99/30.57}                                                                                          & { {49.81/38.21/33.96}}                                                                           & { 32.35/27.88/26.97}                                                                                    & \textbf{ 34.82/33.06/32.13/31.63}                                                                                          & { 34.55/32.97/32.04/31.49}                                                                                                   \\ \hline
		{ {Mushroom}}        & { 29.48/28.11/27.52/27.05}                                                                                          & { 28.95/27.86/27.33/26.98}                                                                                          & { 29.69/28.13/27.38/26.83}                                                                                          & { 35.21/{30.54}/28.47}                                                                                    & { 28.87/25.59/24.62}                                                                                    & { 31.07/29.78/29.18/\textbf{28.75}}                                                                                          & { \textbf{31.26}/\textbf{29.89}/\textbf{29.20}/28.71}                                                                                          \\ \hline
		{ {PenContainer}}    & { 28.80/27.14/26.20/25.56}                                                                                          & { 28.29/26.59/25.96/25.25}                                                                                          & { 22.38/19.93/18.87/18.30}                                                                                          & { 34.46/28.79/25.81}                                                                                    & { 25.94/23.07/22.06}                                                                                    & { 30.81/29.02/28.09/27.42}                                                                                          & { \textbf{31.05/29.23/28.15/27.43}}                                                                                          \\ \hline
		{ {PingpongBat}}     & { 32.29/30.50/29.41/28.77}                                                                                          & { 31.99/29.79/28.78/28.41}                                                                                          & { 29.32/27.18/26.08/25.39}                                                                                          & { 37.46/31.65/28.99}                                                                                    & { 29.60/26.97/26.15}                                                                                    & { 34.37/32.28/31.13/30.30}                                                                                          & { \textbf{34.90/32.65/31.37/30.40}}                                                                                          \\ \hline
		{ {PuerTea}}         & { 29.98/28.40/27.71/27.23}                                                                                          & { 29.78/28.08/27.52/27.00}                                                                                          & { 29.99/28.15/27.32/26.76}                                                                                          & { 35.53/{30.41}/28.30}                                                                                    & { 28.20/26.04/25.22}                                                                                    & { 31.63/30.00/\textbf{29.15/28.70}}                                                                                          & { \textbf{31.88}/\textbf{30.19}/\textbf{29.15}/28.64}                                                                                          \\ \hline
		{ {QQ dog}}          & { 35.26/33.42/32.33/31.61}                                                                                          & { -/-/-/-}                                                                                                          & { 33.88/31.44/30.14/29.24}                                                                                          & { -/-/-/-}                                                                                              & { -/-/-}                                                                                                & { 37.46/35.73/34.77/\textbf{34.12}}                                                                                          & { \textbf{37.86/36.05/34.90}/34.07}                                                                                         \\ \hline
		{ {Salt box}}        & { 25.21/23.53/22.87/22.41}                                                                                          & { 25.33/23.52/22.55/22.27}                                                                                          & { 25.31/23.38/22.45/21.89}                                                                                          & { 30.99/{25.72}/23.48}                                                                                    & { 23.65/21.16/20.23}                                                                                    & { 26.94/25.23/24.32/\textbf{23.74}}                                                                                          & { \textbf{27.17}/\textbf{25.32}/\textbf{24.35}/23.69}                                                                                          \\ \hline
		{ {Shop}}            & { 26.59/24.93/24.40/24.00}                                                                                          & { 23.79/22.48/22.74/22.60}                                                                                          & { 25.08/23.06/21.95/21.39}                                                                                          & { 27.14/24.96/23.14}                                                                                    & { -/-/-}                                                                                                & { 28.45/27.05/26.35/25.87}                                                                                          & { \textbf{28.80/27.27/26.43/25.97}}                                                                                          \\ \hline
		{ {Statue 1}}        & { 32.79/30.79/29.66/28.93}                                                                                          & { 31.70/30.12/28.75/28.55}                                                                                          & { 32.44/29.97/28.79/28.01}                                                                                          & { 38.81/32.20/29.00}                                                                                    & { -/-/-}                                                                                                & { 34.97/32.66/31.51/30.75}                                                                                          & { \textbf{35.53/33.11/31.74/30.84}}                                                                                          \\ \hline
		{ {SweetPotato}}    & { 32.48/31.48/31.07/30.83}                                                                                          & { 32.02/31.24/30.87/30.59}                                                                                          & { 33.45/32.28/31.81/31.51}                                                                                          & { 37.40/{33.94/32.80}}                                                                                    & { 31.42/29.57/28.86}                                                                                    & { 33.87/32.98/32.56/32.36}                                                                                          & { \textbf{34.09/33.03/32.66/32.37}}                                                                                          \\ \hline
		{ {Toy1}}            & { 26.53/25.13/24.56/23.97}                                                                                          & { 23.95/23.10/22.65/22.65}                                                                                          & { 24.14/21.81/20.47/19.59}                                                                                          & { 27.33/24.99/23.13}                                                                                    & { -/-/-}                                                                                                & { 28.06/26.74/26.23/25.77}                                                                                          & { \textbf{28.33/26.95/26.37/25.80}}                                                                                          \\ \hline\hline
		{ {Average}}         & { }                                                                                                                 & { }                                                                                                                 & { }                                                                                                                 & { }                                                                                                     & { }                                                                                                     & { }                                                                                                                 & { }                                                                                                                          \\ \hline
		{ {Average  (Set1)}} & \multirow{-2}{*}{{ \begin{tabular}[c]{@{}c@{}}31.86/30.10/29.24/28.65\\      31.58/29.63/28.82/28.28\end{tabular}}} & \multirow{-2}{*}{{ \begin{tabular}[c]{@{}c@{}}28.29/26.83/26.15/25.80\\      28.84/27.40/26.69/26.32\end{tabular}}} & \multirow{-2}{*}{{ \begin{tabular}[c]{@{}c@{}}29.46/27.24/26.18/25.51\\      29.60/27.40/26.32/25.63\end{tabular}}} & \multirow{-2}{*}{{ \begin{tabular}[c]{@{}c@{}}36.99/31.46/28.67\\      35.87/30.76/28.09\end{tabular}}} & \multirow{-2}{*}{{ \begin{tabular}[c]{@{}c@{}}28.27/25.04/23.86\\      28.79/25.38/23.99\end{tabular}}} & \multirow{-2}{*}{{ \begin{tabular}[c]{@{}c@{}}33.89/32.10/31.17/30.52\\      33.57/31.86/30.95/30.34\end{tabular}}} & \multirow{-2}{*}{{ \textbf{\begin{tabular}[c]{@{}c@{}}34.24/32.36/31.33/30.57\\      33.90/32.10/31.10/30.39\end{tabular}}}} \\ \hline
	\end{tabular}}
	\label{tab:color_psnr}%
\end{table*}%
\subsubsection{Benchmarks and Implementations}
To evaluate the performance of point cloud up-sampling, five state-of-the-art attribute up-sampling schemes were implemented as benchmark schemes and compared with the proposed JGAU methods, which are the MFU \cite{[6]}, FSMMR \cite{[38]}, FGTV \cite{[44]}, {CU-NET \cite{[46]}, and IT-DL-PCC \cite{IT-DL-PCC}. The IT-DL-PCC was retrained based on the pre-trained model by using 28 point clouds from the JPEG Pleno dataset \cite{JPEG-Pleno}. It supports only 2$\times$ and 4$\times$ up-sampling. So, the 8$\times$ up-sampling was calculated by the combination of performing 4$\times$ and 2$\times$ up-samplings. Similarly, the 16$\times$ up-sampling was obtained by using 4$\times$ up-sampling twice. The CU-NET was trained from FaceScape dataset according to \cite{[46]}. It supports 2$\times$, 5$\times$, and 10$\times$ up-sampling. However, unlike IT-DL-PCC and other up-sampling schemes, the CU-NET down-samples source point clouds by its own network and then up-samples them for quality evaluation \cite{[46]}. It requires the original point cloud as the geometrical coordinate support, such that it cannot achieve other up-sampling ratios by performing 2$\times$ up-sampling multiple times.} These schemes were used to up-sample the attribute of sparse point cloud to dense ones.

As for the proposed JGAU, Adam optimizer \cite{[43]} with a learning rate of 0.001 was used to train the geometry and the attribute up-sampling networks. Since the JGAU model is trained at the patch level, we thus generate the patch-level training data from the 75 point clouds. Firstly, {we divided a large-scale point cloud into $N$ patches based on Eq.\eqref{patch}, where the overlap ratio $c$ and $m$ in each patch were set as $300\%$ and 256, respectively.} FPS was used in down-sampling to ensure the selected seed points were uniformly distributed on the point cloud surface. Secondly, $K$-NN was used to find the closest $mR$ points to the seeds, which formed a ground truth patch with $mR$ points, $R\in\{4\times, 8\times, 12\times, 16\times\}$. Finally, we randomly down-sampled the ground truth by a factor of $R$ to form the sparse point clouds. To avoid the over-fitting problem, we not only randomly shuffled the dataset, but also randomly perturbed, scaled, and added Gaussian noise to augment the training dataset. Due to the limited GPU memory capacity, batch size was set to 40 when the up-sampling rate $R$ was 4$\times$, 8$\times$, and 12$\times$. The batch size was set to 28 when the up-sampling rate $R$ was 16$\times$. Moreover, the JGAU has two sub-algorithms in the coarse attribute up-sampling, i.e., GDWAI and DLAI, which are denoted as ``JGAU-GDWAI" and ``JGAU-DLAI", respectively. We trained the proposed JGAU networks on the TensorFlow platform for 400 epochs. All the training and testing experiments were performed on a workstation with 24G RTX3090 GPU. 

\subsubsection{Evaluation Metrics}
We evaluated the up-sampled geometry and attribute information of point clouds with different quantitative evaluation metrics. To evaluate the geometry up-sampling methods, we employed four widely-used geometrical evaluation metrics, which are CD \cite{CD-ICCV}, Hausdorff Distance (HD), Point-to-surface distance (P2F), and Jensen-Shannon Divergence (JSD), to quantitatively measure the quality of up-sampled point clouds. In these metrics, lower value indicates a better up-sampled geometry, and vice versa. To measure the quality of the up-sampled attribute, we used the MPEG software to calculate the attribute PSNR \cite{[47]} between up-sampled and ground truth point clouds. Higher PSNR value indicates a better quality of up-sampled attribute.

\begin{figure*}[h]
	\centering
	\includegraphics[width=0.9\linewidth]{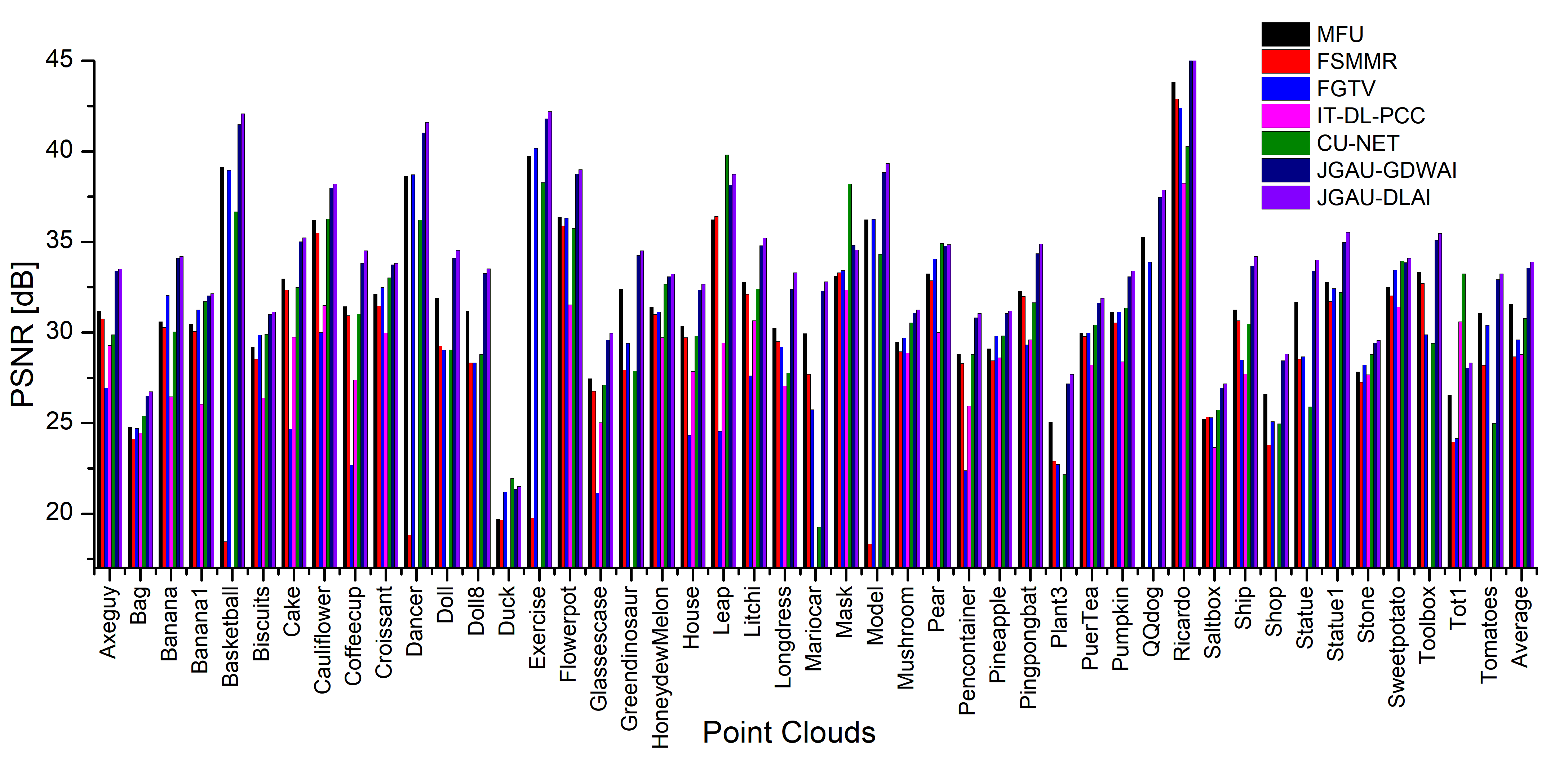}
	\caption{{PSNR comparisons of 4$\times$ point cloud up-sampling using MFU \cite{[6]}, FSMMR \cite{[38]}, FGTV \cite{[44]}, CU-NET (5$\times$)\cite{[46]}, IT-DL-PCC\cite{IT-DL-PCC}, JGAU-GDWAI and JGAU-DLAI.}}
	\label{psnr_4x}
\end{figure*}

\begin{figure*}[t]
	\subfigure[]{
		\begin{minipage}{0.13\linewidth}
			\includegraphics[scale=0.48]
			{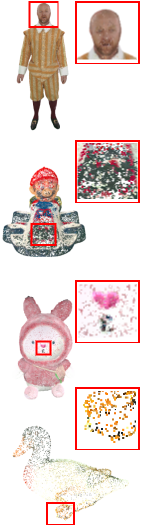}
		\end{minipage}%
	}%
	\subfigure[]{
		\begin{minipage}{0.13\linewidth}
			\includegraphics[scale=0.48]
			{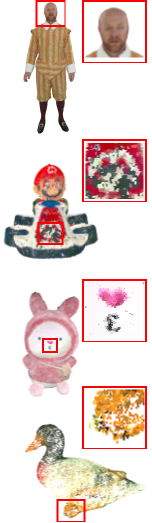}
		\end{minipage}
	}%
	\subfigure[]{
		\begin{minipage}{0.13\linewidth}
			\includegraphics[scale=0.48]
			{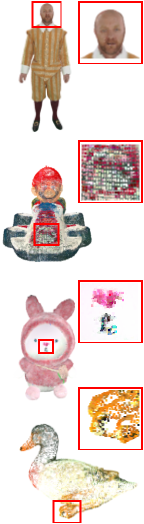}
		\end{minipage}
	}%
	\subfigure[]{
		\begin{minipage}{0.13\linewidth}
			\includegraphics[scale=0.48]
			{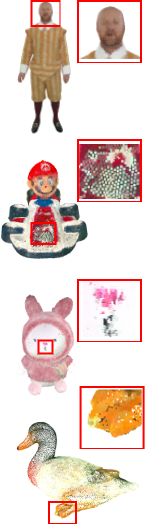}
		\end{minipage}
	}%
	\subfigure[]{
		\begin{minipage}{0.13\linewidth}
			\includegraphics[scale=0.48]
			{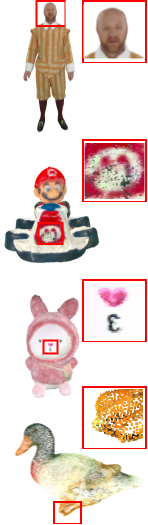}
		\end{minipage}
	}%
	\subfigure[]{
		\begin{minipage}{0.13\linewidth}
			\includegraphics[scale=0.48]
			{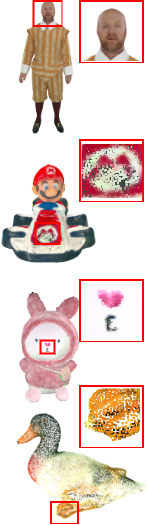}
		\end{minipage}
	}%
	\subfigure[]{
		\begin{minipage}{0.13\linewidth}
			\includegraphics[scale=0.48]
			{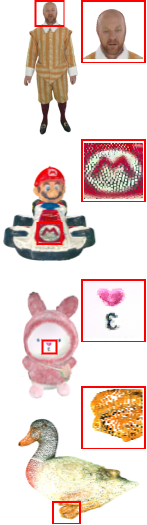}
		\end{minipage}%
	}
	\centering
	\caption{Visualization results of 4$\times$ up-sampled point clouds by five different methods. (a) Source sparse point clouds, (b) MFU, (c) FSMMR, (d) FGTV, (e) JGAU-GDWAI, (f) JGAU-DLAI and (g) Ground truth dense point cloud.}
	\label{fig5}
\end{figure*}

\begin{figure*}[t]
\subfigure[]{
\begin{minipage}{0.13\linewidth}
\includegraphics[scale=0.50]
{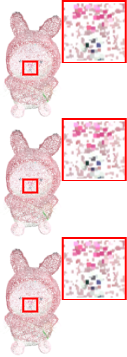}
\end{minipage}
}%
\subfigure[]{
\begin{minipage}{0.13\linewidth}
\includegraphics[scale=0.50]
{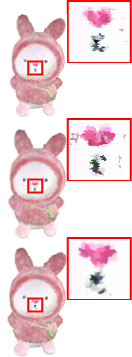}
\end{minipage}
}%
\subfigure[]{
\begin{minipage}{0.13\linewidth}
\includegraphics[scale=0.50]
{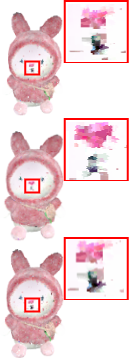}
\end{minipage}
}%
\subfigure[]{
\begin{minipage}{0.13\linewidth}
\includegraphics[scale=0.50]
{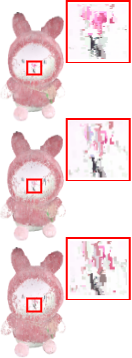}
\end{minipage}
}%
\subfigure[]{
\begin{minipage}{0.13\linewidth}
\includegraphics[scale=0.50]
{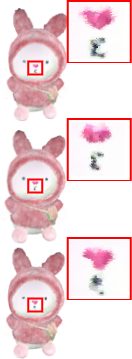}
\end{minipage}
}%
\subfigure[]{
\begin{minipage}{0.13\linewidth}
\includegraphics[scale=0.50]
{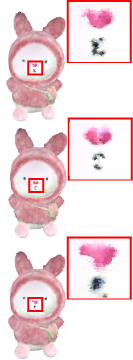}
\end{minipage}
}%
\subfigure[]{
	\begin{minipage}{0.13\linewidth}
		\includegraphics[scale=0.50]
		{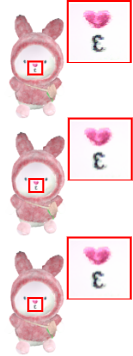}
	\end{minipage}
}%
\centering
\caption{Visualization results of 8$\times$, 12$\times$ and 16$\times$ up-sampled point cloud for \emph{Doll8}. Top row to bottom row are 8$\times$, 12$\times$, and 16$\times$, (a) Source sparse point clouds, (b) MFU, (c) FSMMR, (d) FGTV, (e) JGAU-GDWAI, (f) JGAU-DLAI and (g) Ground truth.}
\label{scale}
\end{figure*}
\subsection{Performance Comparison for Attribute Up-sampling}
Table \ref{tab:color_psnr} shows the PSNR comparisons of the up-sampled point cloud from MFU \cite{[6]}, FSMMR \cite{[38]}, FGTV \cite{[44]}, {CU-NET \cite{[46]}, IT-DL-PCC \cite{IT-DL-PCC}}, JGAU-GDWAI, and JGAU-DLAI, where the values in bold denote the best ones. In the experiment, the up-sampling rate $R$ was set as 4$\times$, 8$\times$, 12$\times$ and 16$\times$, respectively. Note that the symbol ``-" for $QQdog$ is not presented and counted in the average value. The main reason is the buffer overflowed in running FSMMR due to large-scale point clouds. {CU-NET has its own downsampling and cannot process $QQdog$ due to larger geometry bit depth. The IT-DL-PCC scheme can only up-sample point clouds with coordinates that are positive integers, so the point clouds that cannot be computed are also labelled as ``-".} ``Average" presents the average value of the 23 point clouds, half of Set1, in the table and ``Average (Set1)" is the average value of all 46 point clouds in the Set1.

We have three key observations. {First}, when the $R$ was 4$\times$, the average PSNR of the up-sampled point clouds generated by MFU, FSMMR, FGTV and {IT-DL-PCC are 31.86 dB, 28.29 dB, 29.46 dB and 28.27dB}, respectively. The PSNR achieved by the JGAU-GDWAI ranges from 26.94 dB to 41.80 dB, and 33.89 dB on average, which are better than those schemes. Moreover, the PSNR achieved by the JGAU-DLAI ranges from 27.17 dB to 42.20 dB, and 34.24 dB on average, most of which are the best among all schemes. The average PSNR achieved by the JGAU-DLAI is higher than those of the MFU, FSMMR, FGTV and {IT-DL-PCC for 2.38 dB, 5.95 dB, 4.78 dB and 5.97dB,} which are significant gains. Moreover, Fig.\ref{psnr_4x} shows the histogram of the PSNRs of the up-sampled point clouds at rate $4\times$. We can observe that the proposed JGAU-GDWAI and JGAU-DLAI are the highest two schemes among {the seven} schemes and JGAU-DLAI is superior to JGAU-GDWAI.

{Second}, when the up-sampling rate $R$ was 8$\times$, the average PSNRs of JGAU-GDWAI and JGAU-DLAI were 32.10 dB and 32.36 dB, respectively, which are the top two among the benchmark schemes. Similar results can be found when the up-sampling rate $R$ was 12$\times$ and 16$\times$ according to Table \ref{tab:color_psnr}. {We can observe that the average PSNR of CU-NET is 30.76 dB for 5$\times$ up-sampling. In this case, the average PSNR of CU-NET in 8$\times$ up-sampling, if applicable, will be lower than 30.76dB. The average PSNRs achieved by the proposed JGAU-GDWAI and JGAU-DLAI are 31.86dB and 32.10dB in 8$\times$ up-sampling, which are higher than 30.76 dB. There are some outliers. As for the $Mask$, the CU-NET achieves 49.81 dB, 38.21dB and 33.96 dB PSNR for 2$\times$, 5$\times$ and 10$\times$ up-sampling, which are much higher than the rest of schemes. The main reason are that the CU-NET is developed for human face and the $Mask$ is smooth. The CU-NET achieves 41.83 dB and 50.35 dB for the $Leap$ and $Basketball$ at 2$\times$ up-sampling. The main reason is the CU-NET has a low sampling rate and the sparse point cloud is down-sampled by its own network, which is different from other benchmark schemes.} When the up-sampling factor was 16$\times$, the JGAU-DLAI is inferior to JGAU-GDWAI for some point clouds, such as \emph{Mask}, \emph{Banana} and \emph{QQdog} etc. The main reason is that the deep neural network is difficult to extract meaningful high-dimensional attribute features when the local information is too sparse. At that time, the handcrafted distance weighted importance is effective in some cases.

{Third}, in terms of ``Average (Set1)" that has 46 point clouds, JGAU-DLAI achieves an average PSNR of 33.90 dB, 32.10 dB, 31.10 dB, and 30.39 dB when up-sampling rates are 4$\times$, 8$\times$, 12$\times$, and 16$\times$, respectively, which are superior to the MFU, FSMMR, FGTV, and IT-DL-PCC. Compared to the state-of-the-art MFU scheme, the JGAU-DLAI achieves an average of 2.32 dB, 2.47 dB, 2.28 dB and 2.11 dB PSNR gains at four up-sampling rates, respectively, which are significant. Overall, similar results and gains can be found as compared with those in ``Average", which indicates that the JGAU-DLAI and JGAU-GDWAI are stable and generalizable to more and diverse point clouds.

\begin{table}[t]
	\centering
	\caption{Average PSNR of 4$\times$ up-sampled point clouds for cross validation in SYSU-PCUD.[Unit:dB]}
	\begin{tabular}{|c|c|c|c|c|c|c|}
		\hline
	{Training}&{Testing}& \multirow{2}[1]{*}{MFU}&\multirow{2}[1]{*}{FSMMR}&\multirow{2}[1]{*}{FGTV}&\multicolumn{2}{c|}{JGAU}\\
\cline{6-7}
        {Set}&{Set} & & & & GDWAI&DLAI\\
		\hline
Set2,Set3&Set1	&31.58 	&28.84 	&29.60 	&33.57 &	\textbf{33.90}\\
		\hline
Set1,Set3&Set2	&30.46& 27.22	&28.75 	&32.52 	&\textbf{32.88}\\
		\hline
Set1,Set2&Set3	&29.65 & 27.10	&27.74 	&31.63 	&\textbf{32.01}\\
		\hline
\multicolumn{2}{|c|}{Average}	&30.56 	&27.72 	&28.70 	&32.57 	&\textbf{32.93}\\
		\hline
	\end{tabular}%
	\label{tab:cross_validation}%
\end{table}%

\begin{table*}[t]
	\centering
	\caption{{PSNR comparisons of the up-sampled point clouds by MFU \cite{[6]}, FSMMR \cite{[38]}, FGTV \cite{[44]}, CU-NET \cite{[46]}, IT-DL-PCC \cite{IT-DL-PCC}, JGAU-GDWAI and JGAU-DLAI in the JPEG Pleno dataset \cite{JPEG-Pleno}, where the best is in bold.[Unit:dB]}}
	\scalebox{0.9}{\begin{tabular}{|c|c|c|c|c|c|c|}
			\hline
			{ {Schemes}}             & { {MFU}}      & { {FSMMR}} & { {FGTV}}              & { {CU-NET}}    & { {IT-DL-PCC}} & { {JGAU-DLAI}}         \\ \hline
			{ {Point Clouds}}        & { 4$\times$/8$\times$/16$\times$}         & { 4$\times$/8$\times$/16$\times$}      & { 4$\times$/8$\times$/16$\times$}                  & { 2$\times$/5$\times$/10$\times$}         & { 4$\times$/8$\times$/16$\times$}          & { 4$\times$/8$\times$/16$\times$}                  \\ \hline
			{ {PC\_09}}              & { 28.61/27.44/26.87} & { -/-/-}          & { \textbf{30.29/29.27}/28.47} & { 30.31/26.60/24.87} & { 28.31/27.75/26.58}  & { 30.06/29.24/\textbf{28.65}}          \\ \hline
			{ {RWT120}}              & { 31.72/30.36/29.35} & { 29.97/-/-}    & { 32.90/31.34/30.21}          & { 29.58/26.02/23.88} & { 29.92/28.79/27.28}  & { \textbf{33.70/32.27/31.09}}  \\ \hline
			{ {RWT130}}              & { 26.67/24.60/23.53} & { 24.59/-/-}      & { 28.06/26.15/24.60}          & { 27.51/23.24/21.25} & { 23.54/22.78/21.37}  & { \textbf{29.01/27.23/25.64}} \\ \hline
			{ {RWT134}}              & { 28.60/27.14/26.00} & { 26.43/-/-}      & { 29.44/27.78/26.62}          & { 28.07/24.86/23.31} & { 26.04/25.59/24.13}  & { \textbf{30.51/28.81/27.68}} \\ \hline
			{ {RWT136}}              & { 31.67/30.45/29.56} & { 29.97/-/-}      & { 32.32/31.01/29.87}          & { 29.27/26.71/25.11} & { 29.34/28.60/27.41}  & { \textbf{33.90/32.56/31.43}}  \\ \hline
			{ {RWT144}}              & { 28.68/28.47/26.63} & { 28.30/-/-}       & { 31.50/28.89/26.78}          & { 30.93/24.82/21.78} & { 26.49/25.18/22.89}  & { \textbf{33.57/30.79/28.05}} \\ \hline
			{ {RWT152}}              & { 33.16/31.12/29.71} & { -/-/-}          & { 34.31/32.18/30.54}          & { 31.59/26.55/23.70} & { 29.99/29.11/26.78}  & { \textbf{35.46/33.45/31.73}} \\ \hline
			{ {RWT2}}                & { 36.25/34.02/32.14} & { -/-/-}          & { 37.37/34.98/32.95}          & { 34.90/29.18/25.87} & { 32.52/31.40/29.17}  & { \textbf{38.94/36.67/34.57}} \\ \hline
			{ {RWT246}}              & { 30.05/28.42/27.14} & { -/-/-}          & { 30.80/28.87/27.55}          & { 30.10/26.04/24.21} & { 27.88/27.10/25.64}  & { \textbf{32.31/30.27/28.72}} \\ \hline
			{ {RWT305}}              & { 28.80/26.40/24.50} & { 26.90/-/-}       & { 29.89/27.76/26.06}          & { 28.84/23.41/20.87} & { 25.50/24.45/22.84}  & { \textbf{31.36/29.25/27.12}} \\ \hline
			{ {RWT34}}               & { 37.50/35.68/33.90} & { -/-/-}      & { 38.39/35.99/34.12}          & { 35.57/30.10/26.94} & { 33.69/32.43/29.51}  & { \textbf{39.93/37.72/35.67}} \\
\hline
			{ {RWT374}}              & { 41.09/39.02/37.31} & { -/-/-}          & { 41.82/39.41/37.38}          & { 39.78/33.82/30.53} & { 36.38/35.08/31.80}  & { \textbf{43.78/41.49/39.36}} \\ \hline
			{ {RWT395}}              & { 30.49/28.35/26.88} & { -/-/-}          & { 30.19/28.54/27.02}          & { 30.50/25.60/23.42} & { 27.37/26.50/25.49}  & { \textbf{32.92/30.60/28.70}}   \\ \hline
			{ {RWT430}}              & { 33.27/31.33/29.64} & { -/-/-}          & { 34.21/31.76/29.82}          & { 34.64/28.99/26.58} & { 28.95/28.12/26.70}  & { \textbf{36.14/33.53/31.3}}  \\ \hline
			{ {RWT462}}              & { 31.40/29.68/28.35} & { -/-/-}          & { 31.39/29.79/28.55}          & { 31.20/26.79/24.78} & { 28.71/27.88/26.60}  & { \textbf{33.23/31.24/29.86}} \\ \hline
			{ {RWT473}}              & { 30.31/28.32/26.85} & { -/-/-}          & { 31.33/29.02/27.28}          & { 30.55/25.56/23.16} & { 26.83/26.12/24.13}  & { \textbf{32.80/30.48/28.48}}  \\ \hline
			{ {RWT501}}              & { 40.49/38.59/36.97} & { 38.49/-/-}    & { 41.53/39.25/37.35}          & { 38.79/33.23/29.75} & { 36.56/35.16/32.52}  & { \textbf{42.88/41.00/39.15}}    \\ \hline
			{ {RWT503b}}             & { 30.98/29.06/27.53} & { 29.11/-/-}    & { 32.08/29.87/28.19}          & { 29.76/25.15/22.87} & { 27.08/26.46/24.51}  & { \textbf{33.28/31.20/29.20}}   \\ \hline
			{ {RWT529}}              & { 34.79/32.83/31.18} & { 32.90/-/-}      & { 34.10/32.39/31.01}          & { 34.54/29.47/26.72} & { 31.03/30.20/27.50}  & { \textbf{37.41/34.96/32.94}} \\ \hline
			{ {RWT53}}               & { 31.48/29.53/28.05} & { -/-/-}          & { 32.37/30.28/28.64}          & { 30.13/25.44/23.08} & { 28.81/27.88/26.20}  & { \textbf{33.99/31.77/29.87}} \\ \hline
			{ {RWT70}}               & { 28.69/26.61/25.10} & { 26.36/-/-}      & { 29.89/27.66/25.93}          & { 27.77/22.70/20.24} & { 25.99/24.97/23.52}  & { \textbf{31.09/28.89/27.03}} \\ \hline
			{ {RuaDeCoimbra\_vox10}} & { 23.70/22.02/21.05} & { -/-/-}          & { 24.03/22.74/21.50}          & { 23.67/19.71/17.73} & { 22.27/21.23/19.99}  & { \textbf{25.82/24.05/22.72}} \\ \hline
			{ {Goat\_skull}}         & { 29.18/27.65/26.78} & { -/-/-}          & { 30.21/28.72/27.60}          & { 29.38/25.68/23.88} & { 27.23/26.17/24.17}  & { \textbf{31.08/29.74/28.66}} \\ \hline
			{ {Kinfuscene0043}}      & { 26.55/24.58/23.25} & { -/-/-}          & { 26.87/25.11/23.84}          & { 37.78/24.87/21.60} & { 24.69/23.05/21.04}  & { \textbf{28.04/26.26/24.41}} \\ \hline
			{ {Kinfuscene0069}}      & { 31.77/29.72/28.55} & { -/-/-}          & { 32.14/30.41/29.11}          & { 48.13/31.70/28.03} & { 29.04/27.35/25.09}  & { \textbf{33.35/31.46/30.09}} \\ \hline
			{ {Average}}             & { 31.44/29.66/28.27} & { 30.14/-/-}      & { 32.30/30.37/28.84}          & { 32.13/26.65/24.17} & { 28.57/27.57/25.71}  & { \textbf{33.78/31.80/30.14}}  \\ \hline
	\end{tabular}}
	\label{tab:color_psnr_jpeg}%
\end{table*}%

\begin{table*}[t]
	\centering
	\caption{{PSNR comparisons of the up-sampling compressed point cloud by MFU \cite{[6]}, FSMMR \cite{[38]}, FGTV \cite{[44]}, CU-NET \cite{[46]}, IT-DL-PCC \cite{IT-DL-PCC}, JGAU-GDWAI and JGAU-DLAI on compressed 8iVFBv2 datasets, where the best is in bold.[Unit:dB]}}
	\setlength{\tabcolsep}{1mm}
	\scalebox{0.9}{\begin{tabular}{|c|c|c|c|c|c|c|}
			\hline
			{ {Schemes}}                           & { {MFU}}       & { {FSMMR}}     & { {FGTV}}      & { {CU-NET}}     & { {IT-DL-PCC}} & { {JGAU-DLAI}}         \\ \hline
			{ {Point Clouds}}                      & { {4$\times$/8$\times$/16$\times$}} & { {4$\times$/8$\times$/16$\times$}} & { {4$\times$/8$\times$/16$\times$}} & { {2$\times$/5$\times$/10$\times$}} & { {4$\times$/8$\times$/16$\times$}} & { {4$\times$/8$\times$/16$\times$}}         \\ \hline
			{ {Longdress\_vox10\_1051\_rL\_rec}}   & { 28.30/26.93/23.44}  & { 26.94/26.09/24.95}  & { 28.26/27.03/25.79}  & { 28.01/26.49/24.88}  & { 26.24/25.47/23.95}  & { \textbf{29.37/28.39/27.28}} \\ \hline
			{ {Longdress\_vox10\_1051\_rH\_rec}}   & { 29.62/27.34/23.61}  & { 26.01/25.57/23.87}  & { 29.59/27.49/25.79}  & { 27.78/26.52/24.90}  & { 26.49/25.55/23.91}  & { \textbf{31.99/29.78/27.69}} \\ \hline
			{ {Loot\_vox10\_1000\_rL\_rec}}        & { 35.94/35.11/33.95}  & { 32.81/31.87/31.45}  & { 35.91/35.16/34.32}  & { 35.18/34.52/33.26}  & { 34.25/33.33/31.26}  & { \textbf{36.47/36.20/35.69}} \\ \hline
			{ {Loot\_vox10\_1000\_rH\_rec}}        & { 38.90/37.21/33.25}  & { 34.60/33.62/31.67}  & { 38.85/36.83/35.23}  & { 34.54/34.43/33.34}  & { 35.04/33.85/31.49}  & { \textbf{41.19/39.19/37.50}} \\ \hline
			{ {Redandblack\_vox10\_1450\_rL\_rec}} & { 33.20/30.39/27.9}   & { 29.69/31.40/30.60}  & { 33.06/32.27/31.20}  & { 30.07/29.88/29.16}  & { 30.69/29.88/27.21}  & { \textbf{33.95/33.65/32.97}} \\ \hline
			{ {Redandblack\_vox10\_1450\_rH\_rec}} & { 35.34/30.92/28.2}   & { 27.16/26.66/26.19}  & { 35.04/33.24/31.58}  & { 27.53/27.82/27.82}  & { 30.90/30.02/27.32}  & { \textbf{37.87/36.01/34.15}} \\ \hline
			{ {Soldier\_vox10\_0536\_rL\_rec}}     & { 33.27/32.47/31.49}  & { 32.19/31.70/31.07}  & { 33.33/32.48/31.47}  & { 33.10/31.81/30.26}  & { 31.88/30.99/29.58}  & { \textbf{34.04/33.64/32.88}} \\ \hline
			{ {Soldier\_vox10\_0536\_rH\_rec}}     & { 35.88/33.4/31.57}   & { 31.58/31.56/31.68}  & { 35.99/34.06/32.53}  & { 34.55/33.00/31.13}  & { 32.95/31.74/29.93}  & { \textbf{38.31/36.31/34.46}} \\ \hline
			{ {Average}}                           & { 33.81/31.72/29.18}  & { 30.12/29.81/28.93}  & { 33.75/32.32/30.99}  & { 31.35/30.56/29.34}  & { 31.05/30.10/28.08}  & { \textbf{35.40/34.15/32.83}} \\ \hline
	\end{tabular}}
	\label{tab:color_psnr_COMPRESSION}%
\end{table*}%

In addition {to quantitative PSNR evaluation, visual quality of up-sampled point clouds is also compared}. Fig. \ref{fig5} shows the visualization results of the four up-sampled point clouds, including \emph{AxeGuy}, \emph{MarioCar}, \emph{Duck} and \emph{Doll8}, by five up-sampling scheme at 4$\times$ rate. {CU-NET and IT-DL-PCC were not visually compared because the 4$\times$ up-sampling was not supported by CU-NET and some point clouds could not be up-sampled by the IT-DL-PCC.} The rectangles are enlarged parts of the point clouds. It can be observed that the large-scale colored point clouds generated by the proposed JGAU-GDWAI and JGAU-DLAI have fewer artifacts, outliers, cracks, and attribute distortions than those from the three benchmark methods. Moreover, Fig. \ref{scale} shows the visualization results of up-sampled point clouds when $R$ is 8$\times$, 12$\times$ and 16$\times$ for \emph{Doll8}. We observe the proposed JGAU-GDWAI and JGAU-DLAI achieve better visual quality than the benchmarks. Moreover, higher quality and less artifacts are achieved when the sampling rate $R$ is smaller. Overall, from both the statistical PSNR and visualization results, it is found that the proposed JGAU-DLAI is the best and JGAU-GDWAI is the second best at different up-sampling rates, which proves that the proposed JGAU is effective.

\subsection{Cross Validation for Attribute Up-sampling}
To validate the robustness of the JGAU, cross validations were implemented by changing the training and testing sets. The database of 121 source point clouds and their corresponding sparse versions from down-sampling were divided into three sets, which are 46 point clouds in Set1, 38 point clouds in Set2 and the rest 37 point clouds in Set3. In the previous subsection, Set2 and Set3 were used as training and Set1 was used as testing. Alternatively, in this cross validation, Set2 and Set3 were used in testing respectively while the rest point clouds were used as the training sets.

Table \ref{tab:cross_validation} demonstrates the average PSNR of the up-sampled point clouds generated from different training and testing sets when the up-sampling ratio is 4$\times$. Because buffer overflows while using the FSMMR to up-sample point cloud $Rafa$ in Set2, this result is not included in the average PSNR of FSMMR. The average PSNRs of the MFU, FSMMR, and FGTV are 30.56 dB, 27.72 dB, and 28.70 dB, respectively. The MFU is the best among the three schemes. The average PSNRs of the JGAU-GDWAI and JGAU-DLAI are 32.57 dB and 32.93 dB, which are 2.01 dB and 2.37 dB higher than those of MFU. Moreover, JGAU-DLAI is better than the JGAU-GDWAI by 0.36 dB, demonstrating that the learning based coarse up-sampling method is more effective. In addition, the average PSNR of the proposed JGAU-GDWAI and JGAU-DLAI trained from different training and testing sets vary from 31.63 dB to 33.57 dB and from 32.01 dB to 33.90 dB, respectively, which is mainly caused by diverse properties of the testing point clouds. Overall, the proposed JGAU-GDWAI and JGAU-DLAI are stable with different training sets.

\subsection{Cross Database Validation for Attribute Up-sampling}
{To further validate the robustness and applicability of the proposed JGAU, in addition to the SYSU-PCUD, we also tested the proposed JGAU-DLAI and benchmark schemes on point clouds of JPEG Pleno dataset \cite{JPEG-Pleno}, which were used in \cite{IT-DL-PCC}. Three up-sampling rates 4$\times$, 8$\times$, and 16$\times$ were performed for MFU, FSMMSR, FGTV, IT-DL-PCC and JGAU. 2$\times$, 5$\times$, and 10$\times$ up-samplings were performed for CU-NET.
Table \ref{tab:color_psnr_jpeg} shows the PSNR comparisons of the point clouds up-sampling on the JPEG Pleno dataset. The FSMMR method is high complex and encountered memory overflow issues for large point clouds, which were labeled as ``-". We can observe that the average PSNRs of 4$\times$ up-sampling are 31.44 dB, 30.14 dB, 32.30 dB, 28.57 dB and 33.78 dB for MFU, FSMMR, FGTV, IT-DL-PCC and JGAU-DLAI, respectively. The average PSNR of 2$\times$ up-sampling for CU-NET is 32.13 dB, which is lower than the 4$\times$ up-sampling PSNR of the MFU, FGTV and JGAU-DLAI, which indicates the CU-NET is inferior to these schemes and the CU-NET is with low generalization ability. Evidently, the proposed JGAU-DLAI achieves the highest PSNR and outperforms MFU, FGTV, CU-NET, and IT-DL-PCC across all point clouds. Compared to the FGTV, JGAU-DLAI performed better on most point clouds except `$PC\_09$' point cloud for 4$\times$ and 8$\times$ up-sampling. The average PSNR achieved by the JGAU-DLAI are 33.78dB, 31.80 dB and 30.14 dB for 4$\times$, 8$\times$ and 16$\times$ up-sampling, respectively, which are the best among all schemes. The FGTV achieves 32.30 dB, 30.37dB and 28.84 dB, respectively, which are in the second place. These results demonstrate the superiority and strong generalization capability of the proposed JGAU-DLAI.}

{In addition to the cross database validation, the robustness of the proposed JGAU was also evaluated on up-sampling the compressed point clouds. Four point cloud sequences in 8iVFBv2 dataset \cite{8iVFB}, including $Longdress$, $Loot$, $Redandblack$ and $Soldier$, were compressed and decoded by V-PCC\cite{V-PCC,ZhangTIP2023}. The quantization parameters for geometry and attribute coding in V-PCC were set as \{27, 38\} and \{7, 20\} for low and high bitrates, denoted as `rL' and `rH', respectively. Then, the decoded point clouds with compression distortion were up-sampled by using the proposed JGAU and benchmark schemes. Table \ref{tab:color_psnr_COMPRESSION} shows the PSNR comparison between JGAU-DLAI and benchmark schemes in up-sampling the compressed point clouds in 8iVFBv2 dataset. We can observe that the average PSNR achieved by the MFU, FSMMR, FGTV, CU-NET, IT-DL-PCC and JGAU-DLAI are 33.81 dB, 30.12 dB, 33.75 dB, 31.35 dB, 31.05 dB and 35.40 dB, respectively, for 4$\times$ up-sampling. JGAU-DLAI achieves the highest up-sampling quality. Similar results can be found for 8$\times$ and 16$\times$ up-sampling. Overall, the proposed JGAU-DLAI is able to achieve the best up-sampling quality among all schemes and can adapt to the point clouds with high and low compression distortion, which validates the robustness of the proposed JGAU.
}

\begin{table}[t]
	\centering
	\caption{Geometry up-sampling using JGAU and Dis-PU  at different sampling rates $R$.}
	\begin{tabular}{|c|c|c|c|c|c|}
		\hline
		 $R$ & Method &  CD    & HD    & JSD ($10^{-3}$)   & P2F \\
		\hline
		\multirow{2}*{4$\times$} & Dis-PU\cite{[24]}  &  0.53 & 2.30 & 0.85 & 0.12 \\
        \cline{2-6}
		& JGAU &  0.52 & 2.35 & 0.83 & 0.12 \\
        \hline
		\multirow{2}*{8$\times$} & Dis-PU  &  0.70 & 3.26 & 1.23 & 0.16 \\
 \cline{2-6}
		& JGAU &  0.68 & 3.55 & 1.25 & 0.16 \\
        \hline
		\multirow{2}*{12$\times$} & Dis-PU &  0.81 & 3.95 & 1.61 & 0.19 \\
 \cline{2-6}
		& JGAU &  0.81 & 4.05 & 1.55 & 0.19 \\
        \hline
		\multirow{2}*{16$\times$} & Dis-PU &  0.85 & 4.62 & 1.72 & 0.22 \\
 \cline{2-6}
		& JGAU & 0.85 & 4.53 & 1.84 & 0.22 \\
		\hline
	\end{tabular}
	\label{tab:2}
\end{table}
\subsection{Performance Comparison for Geometry Up-sampling}
In addition to the attribute up-sampling, the geometry up-sampling was also evaluated. Since the geometry up-sampling of JGAU originated from Dis-PU \cite{[24]}, we compared it with the Dis-PU, where the geometry quality was measured with CD, HD, JSD, and P2F. Table \ref{tab:2} shows the comparison of the geometry up-sampling performance between JGAU and Dis-PU at different up-sampling rates. We can observe that the average CD of JGAU are 0.52, 0.68, 0.81 and 0.85 for 4$\times$, 8$\times$, 12$\times$ and 16$\times$ up-sampling, respectively. These CD values are slightly lower or the same as those of the Dis-PU, which indicates a slightly better performance owning to the refining with objectives in Eq. \eqref{eq_sub}. Similarly, in terms of HD, JSD and P2F, the geometry quality of the JGAU is similar or slightly better than those of Dis-PU. Overall, the geometry up-sampling of JGAU is generally similar to the Dis-PU.

\begin{table}[t]
	\centering
	\caption{{The computational time for 4$\times$ up-sampling for different methods, where CU-NET is 2 $\times$ up-sampling.}[Unit:S]}
\setlength{\tabcolsep}{0.2mm}
\scalebox{0.85}{	\begin{tabular}{|c|c|c|c|c|c|cc|}
		\hline
		{ }                                                                              & { }                      & { }                        & { }                       & { }                            & { }                        & \multicolumn{2}{c|}{{ JGAU}}                                                     \\ \cline{7-8}
		\multirow{-2}{*}{{ \begin{tabular}[c]{@{}c@{}}Point\\      Clouds\end{tabular}}} & \multirow{-2}{*}{{ MFU}} & \multirow{-2}{*}{{ FSMMR}} & \multirow{-2}{*}{{ FGTV}} & \multirow{-2}{*}{{ IT-DL-PCC}} & \multirow{-2}{*}{{ CU-NET}} & \multicolumn{1}{c|}{{ GDWAI}}           & { DLAI}            \\ \hline
		{ AxeGuy}                                                                        & { 396.16}                & { 4265.73}                 & { 76.11}                  & { 269.21}                      & { 0.120}                   & \multicolumn{1}{c|}{{ 383.04}}          & { 380.78}          \\ \hline
		{ Banana}                                                                        & { 63.00}                    & { 211.66}                  & { 26.64}                  & { 117.26}                      & { 0.005}                   & \multicolumn{1}{c|}{{ 60.06}}           & { 59.13}           \\ \hline
		{ Doll}                                                                          & { 559.54}                & { 958.31}                  & { 101.64}                 & { 125.44}                      & { 0.009}                   & \multicolumn{1}{c|}{{ 544.74}}          & { 540.95}          \\ \hline
		{ Doll8}                                                                         & { 372.4}                 & { 489.1}                   & { 77.27}                  & { 108.15}                      & { 0.007}                   & \multicolumn{1}{c|}{{ 360.42}}          & { 359.86}          \\ \hline
		{ Duck}                                                                          & { 3.52}                  & { 15.39}                   & { 3.84}                   & { 61.28}                       & { 0.003}                   & \multicolumn{1}{c|}{{ 2.75}}            & { 2.66}            \\ \hline
		{ Greendino.}                                                                    & { 21.65}                 & { 38.72}                   & { 17.04}                  & { 25.56}                       & { 0.004}                   & \multicolumn{1}{c|}{{ 18.78}}           & { 18.43}           \\ \hline
		{ Longdress}                                                                     & { 1341.6}                & { 9577.78}                 & { 150.26}                 & { 292.38}                      & { 0.020}                   & \multicolumn{1}{c|}{{ 1317.8}}          & { 1312.68}         \\ \hline
		{ Mariocar}                                                                      & { 127.47}                & { 77.73}                   & { 39.57}                  & { 25.42}                       & { 0.005}                   & \multicolumn{1}{c|}{{ 120.73}}          & { 119.52}          \\ \hline
		{ Plant3}                                                                        & { 143.67}                & { 212.94}                  & { 41.83}                  & { 125.18}                      & { 0.005}                   & \multicolumn{1}{c|}{{ 135.93}}          & { 134.54}          \\ \hline
		{ Ricardo}                                                                       & { 1423.54}               & { 10835.04}                & { 157.43}                 & { 329.33}                      & { 0.020}                   & \multicolumn{1}{c|}{{ 1399.7}}          & { 1391.94}         \\ \hline
		{ Shop}                                                                          & { 214.66}                & { 354.97}                  & { 54.02}                  & { 70.74}                       & { 0.006}                   & \multicolumn{1}{c|}{{ 205.43}}          & { 203.85}          \\ \hline
		{ Statue}                                                                        & { 22.13}                 & { 36.97}                   & { 15.91}                  & { 13.55}                       & { 0.004}                   & \multicolumn{1}{c|}{{ 19.43}}           & { 18.9}            \\ \hline
		{ Tomatoes}                                                                      & { 36.39}                 & { 74.48}                   & { 23.94}                  & { 32.01}                       & { 0.004}                   & \multicolumn{1}{c|}{{ 37.32}}           & { 36.63}           \\ \hline
		{ Toy1}                                                                          & { 372.69}                & { 580.8}                   & { 76.14}                  & { 107.19}                      & { 0.007}                   & \multicolumn{1}{c|}{{ 360.1}}           & { 357.51}          \\ \hline
		{ \textbf{Average}}                                                              & { \textbf{364.17}}       & { \textbf{1980.69}}        & { \textbf{61.55}}         & { \textbf{121.62}}             & { \textbf{0.016}}          & \multicolumn{1}{c|}{{ \textbf{354.73}}} & { \textbf{352.67}} \\ \hline
	\end{tabular}}
	\label{tab:Complexity}%
\end{table}

\subsection{Computational Complexity Analysis}
\label{Sec:CCA}
Computational complexity of the proposed JGAU is also evaluated. The up-sampling model was trained from Set2 and Set3. For simplicity, we randomly selected 14 point clouds from Set1 to analyze the computational time of using the up-sampling models. The up-sampling rate $R$ is 2$\times$ {for CU-NET while $R$ is 4$\times$ for the rest up-sampling methods}. Table \ref{tab:Complexity} shows the comparison of time complexities between the proposed JGAU and the benchmark schemes. We can have the following three observations. Firstly, the computing time of MFU varies significantly from 3.52s to 1341.60s. Similar variations can be found for the FSMMR, FGTV, {IT-DL-PCC and} JGAU schemes, owing to the significant differences in the number of point clouds. Secondly, the average computing time of MFU, FSMMR, FGTV and IT-DL-PCC are 364.17s, 1980.69s, 61.55s and 121.62s, respectively. The average computing time of JGAU-GWDAI and JGAU-DLAI are 354.73s and 352.67s, respectively, which are lower than MFU and FSMMR. {JGAU-DLAI has less computing time as compared with the JGAU-GDWAI due to GPU acceleration and efficient code implementation.} FGTV is with low complexity because it uses a simple interpolation with parameter-free linear program. {In addition, the average computing time of CU-NET is  0.016s, which is the lowest among all up-sampling schemes. This is because it uses sparse convolution as the basic computational unit for feature extraction.} The FSMMR iteratively learns the frequency kernels for mesh-to-mesh resampling, which is highly complex.

\subsection{Ablation Studies}

The parameters and modules of JGAU, i.e. JGAU-GWDAI, are experimentally analyzed, which include the overlapping ratio, the neighborhood size and effectiveness of the AEM.

\subsubsection{Overlapping Ratio}

\begin{table}[t]
	\centering
	\caption{{PSNR and computing time of JGAU-GDWAI using different overlap ratios.}}
	\scalebox{0.85}{
		\begin{tabular}{|c|c|c|c|c|c|}
			\hline
\multicolumn{6}{|c|}{{PSNR (dB)}}\\
			\hline
			Point Clouds & {$c$=1} & {$c$=2}     & {$c$=3} & {$c$=4} & {$c$=5} \\ \hline
			Banana                    & 32.85      & 34.14         & 34.18     & 34.19    & 34.21      \\ \hline
			Doll8                   & 31.95      & 33.61         & 33.51    & 33.40   & 33.27   \\ \hline
			Duck                      & 20.68      & 21.56           & 21.51      & 21.51      & 21.48      \\ \hline
			GreenDinosaur           & 32.21       & 34.48          & 34.51      & 34.47     & 34.44      \\ \hline
			Statue                    & 29.85      & 33.89         & 34.00      & 33.93    & 33.90     \\ \hline
			\textbf{Average}          & \textbf{29.54}      & \textbf{31.54} & \textbf{31.52}      & \textbf{31.50}     & \textbf{31.46}    \\ \hline
			\hline
\multicolumn{6}{|c|}{{Time (s)}}\\
			\hline
			Point Clouds & {$c$=1} & {$c$=2}     & {$c$=3} & {$c$=4} & {$c$=5} \\ \hline
			Banana                    & 12.12      & 24.48          & 56.88      & 77.68      & 98.31      \\ \hline
			Doll8                   & 93.28      & 236.83         & 353.88     & 475.17     & 606.85     \\ \hline
			Duck                      & 0.79       & 1.49           & 2.25       & 3.08       & 4.16       \\ \hline
			GreenDinosaur           & 5.50        & 11.18          & 16.87      & 23.42      & 42.03      \\ \hline
			Statue                    & 5.73       & 11.35          & 17.93      & 24.11      & 43.21      \\ \hline
			\textbf{Average}          & \textbf{23.48}      & \textbf{57.07} & \textbf{89.45}      & \textbf{120.69}     & \textbf{158.91}     \\ \hline
	\end{tabular}}
	\label{tab:7}%
\end{table}%
{We evaluated the up-sampling performance and time complexity at different overlap ratios $c$. We tested the JGAU-GDWAI without overlapping ($c$ = 1) and overlapping with ratios of 200\%, 300\%, 400\%, and 500\%. Also, five sequences, including $Banana$, $Doll8$, $Duck $, $Green\_dinosaur$ and $Statue$, were up-sampled with 4$\times$ up-sampling ratio.
As shown in upper part of the Table \ref{tab:7}, the average PSNR of up-sampled point clouds firstly increases from 29.54 dB to 31.54 dB when $c$ changes from 1 to 2. It makes sense since the boundaries of the patches are significantly improved by using the overlapped patches. However, as the overlap ratio increases from 2 to 5, the average PSNR slightly reduces from 31.54 dB to 31.46 dB. This is because the overlapped regions are up-sampled from different source points, which cause contradictions reducing the PSNR. For the $Banana$, the PSNR improves consistently as the overlap ratio increases, which is mainly because $Banana$ has a simple geometric structure and uniform color distribution. Overall, the PSNR reaches the peak when $c$ is 2 or 3. In addition, the bottom part of the Table \ref{tab:7} shows the time complexity when $c$ is from 1 to 5. We can observe the time complexity increases linearly as the $c$ increases. So, in this paper, $c$ is set as 3.}

\subsubsection{Number of Local Neighborhood Points in AEM}

We selected 14 point clouds from Set1 as testing set of this ablation study, which is the same as settings in subsec. \ref{Sec:CCA}.
The proposed JGAU is highly dependent on the local neighborhood information, which are extracted by using the $K$-NN clustering. In order to analyze the optimal local neighborhood size, we tested $k_2$$\in\{4,8,16,24,32,40\}$ in the AEM of the JGAU-GWDAI. Table \ref{tab:Neighborhood} shows the up-sampled attribute PSNR of different local neighborhood sizes, where the best values are in bold. The up-sampling rate $R$ is 4. The average PSNR of the up-sampled point clouds were 31.59 dB, 31.99 dB, 32.19 dB, 32.18 dB, 32.22 dB and 32.19 dB when $k_2$ is 4, 8, 16, 24, 32, 40, respectively. The JGAU-GWDAI achieves its best when $k_2$ is 32, which is also applicable to the JGAU-DLAI, other sampling rates and the rest point clouds.
\begin{table}[t]
	\centering
	\caption{PSNR of Different Local Neighborhoods in AEM.[Unit:dB]}
	\begin{tabular}{|c|c|c|c|c|c|c|}
		\hline
		$k_2$ & 4    & 8   & 16   & 24  & 32 & 40\\
		\hline
		JGAU-GWDAI &31.59  &31.99 &32.19  &32.18  &\textbf{32.22}  &32.19 \\
\hline

	\end{tabular}%
	\label{tab:Neighborhood}%
\end{table}%
\begin{table}[t]
  \centering
  \caption{PSNR comparison of using AEM.[Unit:dB]}
    \begin{tabular}{|c|c|c|c|c|}
    \hline
     Method & 4$\times$    & 8$\times$   & 12$\times$   & 16$\times$ \\
    \hline
    JGAU-GWDAI w/o AEM & 32.82  & 31.13 & 30.25 & 29.67 \\
    \hline
    JGAU-GWDAI with AEM & \textbf{33.57} & \textbf{31.86} & \textbf{30.95} & \textbf{30.34} \\
    \hline
    \end{tabular}%
  \label{tab:6}%
\end{table}%

\subsubsection{Effectiveness of the AEM}
We also performed an ablation experiment to evaluate the effectiveness of the AEM in the JGAU. We selected Set1 as testing set. Four up-sampling rates were evaluated, i.e., $R\in\{4\times,8\times,12\times,16\times\}$. Table \ref{tab:6} shows the PSNR of the JGAU-GWDAI using AEM and without using AEM, where the best ones are in bold. We can observe that when using the AEM, the average PSNR achieved by the JGAU-GWDAI are 33.57 dB, 31.86 dB, 30.95 dB, and 30.34 dB when $R$ is  4, 8, 12, and 16, respectively, which are all superior to the JGAU-GWDAI without using AEM by 0.75 dB, 0.73 dB, 0.70 dB, and 0.67 dB. These results demonstrate the advantages of using the proposed AEM, which is stable and applicable to various point clouds and up-sampling rates.

\section{Conclusions}
In this paper, we proposed a deep learning-based Joint Geometry and Attribute Up-sampling (JGAU) method to generate large-scale colored point clouds from sparse colored point clouds.
Firstly, we established a large-scale dataset for colored point cloud up-sampling, which has 121 large-scale colored point clouds in six categories and four sampling rates. Secondly, we proposed the framework of the deep learning based JGAU to up-sample the geometry and attribute jointly, where the attribute up-sampling network learns the neighborhood correlation of attribute and geometry distance based importance. Thirdly, we proposed an attribute enhancement module to refine the up-sampled attribute. Extensive experimental results validate that the proposed JGAU is able to achieve an average of 2.11 dB to 2.47 dB PSNR gain, which is significant, as compared with the state-of-the-art. Cross-validation, complexity analysis and ablation studies further reveal that the proposed JGAU is effective and robust.

\bibliographystyle{IEEEtran}

\end{document}